\RequirePackage{fix-cm}
\documentclass[smallextended]{svjour3} 
\smartqed 
\usepackage{graphicx}
\usepackage{subfigure}
\usepackage{algorithm}
\usepackage{algorithmic}

\usepackage{booktabs}
\usepackage{microtype}
\usepackage[hyphens]{url}
\usepackage{amsfonts}
\usepackage{amsmath}
\DeclareMathOperator*{\argmax}{argmax}
\usepackage[misc, geometry]{ifsym}
\usepackage{natbib}
\usepackage{hyperref}

\usepackage{verbatim}
\usepackage{rotating}
\begin{document}

\title{A Framework for Deep Constrained Clustering
}
\subtitle{}


\author{Hongjing Zhang \and
Tianyang Zhan \and Sugato Basu \and Ian Davidson
}


\institute{H. Zhang \at
Department of Computer Science, University of California, Davis. Davis, CA 95616, USA. \\
\email{hjzzhang@ucdavis.edu} 
\and
T. Zhan \at
Department of Computer Science, University of California, Davis. Davis, CA 95616, USA. \\
\email{tzhan@ucdavis.edu} 
\and
S. Basu \at
Google Research, Mountain View, CA, 94043, USA. \\
\email{sugato@google.com}
\and
I. Davidson \at
Department of Computer Science, University of California, Davis. Davis, CA 95616, USA. \\
\email{davidson@cs.ucdavis.edu} 
}

\date{Received: date / Accepted: date}

\maketitle

\begin{abstract}
The area of constrained clustering has been extensively explored by researchers and used by practitioners. Constrained clustering formulations exist for popular algorithms such as k-means, mixture models, and spectral clustering but have several limitations. A fundamental strength of deep learning is its flexibility, and here we explore a deep learning framework for constrained clustering and in particular explore how it can extend the field of constrained clustering. We show that our framework can not only handle standard together/apart constraints (without the well documented negative effects reported earlier) generated from labeled side information but more complex constraints generated from new types of side information such as continuous values and high-level domain knowledge. Furthermore, we propose an efficient training paradigm that is generally applicable to these four types of constraints. We validate the effectiveness of our approach by empirical results on both image and text datasets. We also study the robustness of our framework when learning with noisy constraints and show how different components of our framework contribute to the final performance. Our source code is available at: \url{http://github.com/blueocean92}.

\end{abstract}

\section{Introduction}
\label{intro}

Constrained clustering has a long history in machine learning with many standard algorithms being adapted to be constrained \citep{basu2008constrained} including Expectation-maximization (EM) \citep{basu2004probabilistic}, K-Means \citep{wagstaff2001constrained} and spectral methods \citep{wang2010flexible}. The addition of constraints generated from ground truth labels allows a semi-supervised setting to increase accuracy \citep{wagstaff2001constrained} when measured against the ground truth labeling.

However, there are several limitations in these methods, and one purpose of this paper is to explore how deep learning can make advances to the field beyond what other methods have. In particular, we find that existing non-deep formulations of constrained clustering have the following four limitations:

\emph{Limited Constraints and Side Information}. Constraints are limited to simple together/apart constraints typically generated from labels. However, in some domains, experts may more naturally give guidance at the cluster level, generate constraints from continuous side-information or even complex sources such as ontologies. To address these deficiencies fundamentally new types of constraints are required.

\emph{Negative Effect of Constraints.} For some algorithms though constraints improve performance when \emph{averaged} over many constraint sets, \emph{individual} constraint sets produce results worse than using no constraints \citep{davidson2006measuring} as reported in our earlier paper. However, as a practitioner typically has only one constraint set, constrained-clustering use can be ``hit or miss".

\emph{Intractability and Scalability Issues.} Iterative algorithms that directly solve for clustering assignments run into problems of intractability \citep{davidson2007intractability}. Relaxed formulations (i.e. spectral methods \citep{lu2008constrained,wang2010flexible}) require solving a full rank eigendecomposition problem which takes $O(n^3)$. The deep learning paradigm has shown to be scalable for large data sets and we explore if this is the case for deep constrained clustering.

\emph{Assumption of Good Features.} A critical requirement for existing constrained clustering algorithms is the need for good features or a similarity function. The end-to-end learning benefits of deep learning will be explored to determine if they are useful for constrained clustering.

Though deep clustering with constraints has many potential benefits to overcome these limitations, it is not without its challenges. Our major contributions in this paper are summarized as follows:
\begin{itemize}
\item We propose a deep constrained clustering formulation that can not only encode standard together/apart constraints but a range of new constraint types. Example types include triplet constraints, instance difficulty constraints, and cluster-level balancing constraints (see section \ref{sec:methods}).
\item In addition to generating constraints from instances' labels, we show how our framework can take advantage of continuous side information and an ontology graph to generate triplet constraints and how to learn from multiple constraints simultaneously (see Section \ref{sec:multiple_constraints}).
\item Deep constrained clustering overcomes a long term issue we reported earlier \citep{davidson2006measuring} with constrained clustering of significant practical implications: overcoming the negative effects of individual constraint sets.
\item We show how the benefits of deep learning such as scalability and end-to-end learning translate to our deep constrained clustering formulation.
\item Our method outperforms standard non-deep constrained clustering methods even though these methods are given the auto-encoder embedding provided to our approach (see Table \ref{tab:pairwise_neg}).
\item We show the robustness of our proposed framework (see Section \ref{exp:robustness}) and demonstrate the scalability of our framework on large-scale data set (see Section \ref{exp:runtime_study}).
\item We conduct ablation study and analyze the contributions of each component within our algorithm (see Section \ref{exp:ablation_study}).
\end{itemize}

This paper is an extension of our previous work \citep{zhang2019framework} with the following additions: 1) we show our framework can not only work with constraints generated from ground truth labels but also work with constraints that are generated from an ontology graph which is a weaker form of guidance; 2) whereas previously we conducted deep constrained clustering with only one type of constraints at each run in previous work, in this paper we extend our algorithm to learn with pairwise and triplet constraints simultaneously; 3) we experimentally visualize the learning process of our framework and show how our framework overcomes the negative effects of constraints; 
4) we analyze the effects of noisy constraints on our framework and show the robustness of our model; 5) we analyze each component's contributions within our framework (i.e., initialization, clustering module, constraints learning module) using an ablation study.

The rest of the paper is organized as follows: First, we introduce the related work in section \ref{sec:related work}. We then propose four forms of constraints in section \ref{sec:methods} and introduce how to train the clustering network with these constraints in section \ref{sec:training}. We then discuss the new way of generating constraints and how to learn multiple types of constraints together in section \ref{sec:multiple_constraints}. In section \ref{sec:experiments}, we compare our approach to previous baselines and demonstrate the effectiveness of new types of constraints and also perform a detailed analysis of our proposed framework. Next, we discuss the limitations of current work and conclude in section \ref{sec:conclusion}.

\section{Related Work}
\label{sec:related work}
\textbf{Constrained clustering.} Constrained clustering is an important area, and there is a large body of work that shows how \emph{side information} can improve the clustering performance \citep{wagstaff2000clustering,wagstaff2001constrained,xing2003distance,bilenko2004integrating,wang2010flexible}. Here the side information is typically labeled data which is used to generate \emph{pairwise} together/apart constraints used to partially reveal the ground truth clustering to help the clustering algorithm. Such constraints are easy to encode in matrices and enforce in procedural algorithms though not with its challenges. In particular, we showed \citep{davidson2006measuring} clustering performance improves with larger constraint sets when \textbf{averaged} over many constraint sets generated from the ground truth labeling. However, for a significant fraction (just not the majority) of these constraint sets, the clustering performance is \emph{worse} than using no constraint set. We recreated some of these results in Table \ref{tab:pairwise_neg}.

Moreover, side information can exist in different forms beyond labels (i.e., continuous data), and domain experts can provide guidance beyond pairwise constraints. Some work in the supervised classification setting \citep{joachims2002optimizing,schultz2004learning,schroff2015facenet,gress2016probabilistic} seek alternatives such as relative/triplet guidance, but to our knowledge, such information has not been explored in the non-hierarchical clustering setting. Complex constraints for hierarchical clustering have been explored \citep{bade2008creating,chatziafratis2018hierarchical} but these are tightly limited to the hierarchical structure (i.e., $x$ must be joined with $y$ before $z$) and not directly translated to non-hierarchical (partitional) clustering.

\textbf{Deep Clustering.} Motivated by the success of deep neural networks in supervised learning, unsupervised deep learning approaches are now being explored \citep{xie2016unsupervised,jiang2017variational,yang2017towards,guo2017improved,hu2017learning,ghasedi2017deep,caron2018deep,haeusser2018associative,aljalbout2018clustering,shaham2018spectralnet,ji2019invariant,han2019learning}. There are approaches \citep{yang2017towards,hu2017learning,caron2018deep,shaham2018spectralnet} which learn an encoding that is suitable for a clustering objective first and then applied an external clustering method. Our work builds upon the most direct setting \citep{xie2016unsupervised,guo2017improved} which encodes one self-training objective and finds the clustering allocations for all instances within one neural network.

\textbf{Deep Clustering with Pairwise Constraints.} Most recently, the semi-supervised clustering networks with pairwise constraints have been explored: \citep{hsu2015neural} uses pairwise constraints to enforce small divergence between similar pairs while increasing the divergence between dissimilar pairs assignment probability distributions. However, this approach did not leverage the unlabeled data, hence requires lots of labeled data to achieve good results. Fogel et al. proposed an unsupervised clustering network \citep{fogel2019clustering} by self-generating pairwise constraints from the mutual KNN graph and extends it to semi-supervised clustering by using labeled connections queried from the human. However, this method cannot make out-of-sample predictions and requires user-defined parameters for generating constraints from the mutual KNN graph.

\section{Our Deep Constrained Clustering Framework}
\label{sec:methods}
Here we outline our proposed framework for deep constrained clustering. Our method of adding constraints to and training deep learning can be used for deep clustering methods so long as the network has a $k$ unit output indicating the degree of cluster membership. Here we choose the popular deep embedded clustering method (DEC \citep{xie2016unsupervised}). We sketch this method first for completeness.

\subsection{Deep Embedded Clustering}
We choose to apply our constraints formulation to the deep embedded clustering method DEC \citep{xie2016unsupervised}, which starts with pre-training an autoencoder ($x_i=g(f(x_i)$) but then removes the decoder. The remaining encoder ($z_i=f(x_i)$) is then fine-tuned by optimizing an objective which takes first $z_i$ and converts it to a soft allocation vector of length $k$ which we term $q_{i,j}$ indicating the degree of belief instance $i$ belongs to cluster $j$. Then $q$ is self-trained to produce $p$ a unimodal ``hard'' allocation vector which allocates the instance to primarily only one cluster. We now overview each step.

\textbf{Conversion of $z$ to Soft Cluster Allocation Vector $q$.} Here DEC takes the similarity between an embedded point $z_i$ and the cluster centroid $u_j$ measured by Student's $t$-distribution \citep{maaten2008visualizing}. Note that $v$ is a constant as $v = 1$ and $q_{ij}$ is a soft assignment:
\begin{equation}
\label{dec_q}
q_{ij} = \frac{{(1 + {||z_i - {\mu}_{j}||}^{2} / v)} ^ {-\frac{v+1}{2}}} {\sum_{j^{'}} {(1 + {|| z_i - {\mu}_{j^{'}} ||}^{2} / v)} ^ {-\frac{v+1}{2}}}
\end{equation}

\textbf{Conversion of $Q$ To Hard Cluster Assignments $P$.} The above normalized similarities between embedded points and centroids can be considered as soft cluster assignments $Q$. However, we desire a target distribution $P$ that better resembles a hard allocation vector, $p_{ij}$ is defined as:
\begin{equation}
\label{dec_p}
p_{ij} = \frac{ {q_{ij}}^{2} / \sum_{i} q_{ij} } { \sum_{j^{'}} ({q_{ij^{'}}}^{2} / \sum_{i} q_{ij^{'}})}
\end{equation}

\textbf{Loss Function.} Then, the algorithm's loss function is to minimize the distance between $P$ and $Q$ as follows. Note this is a form of self-training as we are trying to teach the network to produce unimodal cluster allocation vectors.

\begin{equation}
\label{dec_obj}
\ell_{C} = KL(P || Q) = \sum_{i} \sum_{j} p_{ij} \log{\frac{p_{ij}}{q_{ij}}}
\end{equation}

The DEC method requires the initial centroids given ($\mu$) to calculate $Q$ are ``representative". The initial centroids are set using k-means clustering. However, there is no guarantee that the clustering results over an auto-encoders embedding yield a good clustering. We believe that constraints can help overcome this issue which we test later.

\subsection{Different Types of Constraints}
To enhance the clustering performance and allow for more types of interactions between human and clustering models, we propose four types of guidance which are pairwise constraints, instance difficulty constraints, triplet constraints, cardinality, and give examples of each. As traditional constrained clustering methods put constraints on the final clustering assignments, our proposed approach constrains the $q$ vector which is the soft assignment. A core challenge when adding constraints is to allow the resultant loss function to be differentiable so we can derive backpropagation updates.
\subsubsection{Pairwise Constraints}
Pairwise constraints (must-link and cannot-link) are well studied \citep{basu2008constrained} and we showed they are capable of defining any ground truth set partitions \citep{davidson2007intractability}. Here we show how these pairwise constraints can be added to a deep learning algorithm.
We encode the loss for must-link constraints set ML as:
\begin{equation}
\label{must_link_loss}
\ell_{ML} = -\sum_{(a,b) \in ML} \log{\sum_{j} q_{aj} * q_{bj}}
\end{equation}
Similarly loss for cannot-link constraints set CL is:
\begin{equation}
\label{cannot_link_loss}
\ell_{CL} = -\sum_{(a,b) \in CL} \log{(1 - \sum_{j} q_{aj} * q_{bj})}
\end{equation}
Intuitively speaking, the must-link loss prefers instances with same soft assignments and the cannot-link loss prefers the opposite cases.

\subsubsection{Instance Difficulty Constraints}
A challenge with self-learning in deep learning is that if the initial centroids are incorrect, the self-training can lead to poor results. Here we use constraints to overcome this by allowing the user to specify which instances are easier to cluster (i.e., they belong strongly to only one cluster) and ignoring difficult instances (i.e., those that belong to multiple clusters strongly).

We encode user supervision with an $n \times 1$ constraint vector $M$. Let $M_i \in [-1, 1]$ be an instance difficulty indicator, $M_i > 0$ means the instance $i$ is easy to cluster, $M_i = 0$ means no difficulty information is provided and $M_i < 0$ means instance $i$ is hard to cluster.
The loss function is formulated as:
\begin{equation}
\label{instance_loss}
\ell_{I} = \sum_{t \in {\{M_t < 0\}}} - M_t \sum_{j} {q_{tj}}^2 - \sum_{s \in {\{M_s > 0\}}} M_s\sum_{j} {q_{sj}}^2
\end{equation}
The instance difficulty loss function aims to encourage the easier instances to have sparse clustering assignments but prevents the difficult instances having sparse clustering assignments. The absolute value of $M_i$ indicates the degree of confidence in difficulty estimation. This loss will help the model training process converge faster on easier instances and increase our model's robustness towards difficult instances.

\subsubsection{Triplet Constraints}
Although pairwise constraints are capable of defining any ground truth set partitions from labeled data \citep{davidson2007intractability}, in many domains, no labeled side information exists or strong pairwise guidance is not available. Thus we seek triplet constraints, which are weaker constraints that indicate the relationship within a triple of instances.
Given an anchor instance $a$, positive instance $p$ and negative instance $n$ we say that instance $a$ is more similar to $p$ than to $n$. The loss function for all triplets $(a, p, n) \in T$ can be represented as:
\begin{equation}
\label{triplet_loss}
\ell_{T} = \sum_{(a, p, n) \in T}\max (d(q_a, q_n) - d(q_a, q_p) + \theta, 0) 
\end{equation}
where $d(q_a, q_b) = \sum_j q_{aj} * q_{bj}$ and $\theta > 0$. The larger value of $d(q_a, q_b)$ represents larger similarity between $a$ and $b$. The variable $\theta$ controls the gap distance between positive and negative instances. $\ell_{T}$ works by pushing the positive instance's assignment closer to anchor's assignment and preventing negative instance's assignment being closer to anchor's assignment.

\subsubsection{Global Size Constraints}
Experts may more naturally give guidance at a cluster level, previous work \citep{ghasedi2017deep} explored adding uniform distribution assumption to regularize the clustering model. Here we explore clustering size constraints in our framework, which means each cluster should be approximately the same size. Denote the total number of clusters as $k$, total training instances number as $n$, the global size constraints loss function is:
\begin{equation}
\label{global_loss}
\ell_{G} = \sum_{c \in {\{1, .. k\}}}(\sum_{i = 1}^{n} q_{ic}/n - \frac{1}{k})^2
\end{equation}
Our global constraints loss function works by minimizing the distance between the expected cluster size and the actual cluster size. The actual cluster size is calculated by averaging the soft-assignments. To guarantee the effectiveness of global size constraints, we need to assume that the batch size should be large enough to calculate the cluster sizes during our mini-batch training. A similar loss function can be used (see section \ref{sec:extensions}) to enforce other cardinality constraints on the cluster composition such as upper and lower bounds on the number of people with a certain property.

\subsection{Preventing Trivial Solution}
\label{sec:reconstruction}
In our framework the proposed must-link constraints we mentioned before can lead to trivial solution that all the instances are mapped to the same cluster. Previous deep clustering method \citep{yang2017towards} have also met this problem. To mitigate this problem, we combine the reconstruction loss with the must-link loss to learn together. Denote the encoding network as $f(x)$ and decoding network as $g(x)$, the reconstruction loss for instance $x_i$ is:
\begin{equation}
\label{reconstruction_loss}
\ell_{R} = \ell (g(f(x_i)), x_i)
\end{equation}
where $\ell$ is the least-square loss: $\ell(x, y) = {||x - y||}^2$.

\subsection{Extensions to High-level Domain Knowledge-Based Constraints}
\label{sec:extensions}
The constraints proposed in the previous section are typically generated from instance labels or comparisons. A benefit of our framework is the ability to include more complex constraints and we now describe examples of constraints for higher-level domain knowledge.

\textbf{Cardinality Constraints For Fairness.} Cardinality constraints \citep{Dao2016AFF} allow expressing requirements on the number of instances that satisfy some conditions in each cluster. Assume we have $n$ people and want to split them into $k$ groups but wish to minimize disparate impact with respect to gender. Then an example cardinality constraint is to enforce each group should have the same number of males and females. We assume each instance has a protected status variable (PSV) which we called $P$. Then the cardinality constraints can be formulated as:
\begin{equation}
\label{cardinality_loss}
\ell_{Cardinality} = \sum_{c \in {\{1, .. k\}}}(\sum_{P_i = M} q_{ic}/n - \sum_{P_j = F} q_{jc}/n)^2
\end{equation}

For upper-bound and lower-bound based cardinality constraints \citep{Dao2016AFF}, we use the same setting as previously described, now the constraint changes as for each party group we need the number of males to range from $L$ to $U$. Then we can formulate this as:
\begin{equation}
\label{cardinality_bound_loss}
\ell_{CardinalityBound} = \sum_{c \in {\{1, .. k\}}} ({\min(0,\sum_{P_i = M} q_{ic} - L)}^2 + {\max(0,\sum_{P_i = M} q_{ic} - U)}^2)
\end{equation}

\textbf{Logical Combinations of Constraints via Dynamic Addition.} Apart from cardinality constraints, complex logic constraints can also be used to enhance the expressive power of existing constraints. For example, if two instances $x_a$ and $x_b$ are in the same cluster then instances $x_i$ and $x_j$ must be in different clusters (\texttt{Together($x_a,x_b$) $\rightarrow$ Apart($x_i,x_j$)}). This can be achieved in our framework as we can dynamically adding cannot-link constraint $CL(x_i, x_j)$ once we check the soft assignment $q$ of $x_a$ and $x_b$.

Consider a Horn form constraint such as $r \wedge s \wedge t \rightarrow u$. Denote $r = ML(x_a, x_b)$, $s = ML(x_c, x_d)$, $t = ML(x_e, x_f)$ and $u = CL(x_g, x_h)$. By forward passing the instances within $r, s, t$ to our deep constrained clustering model, we can obtain the soft assignment values of these instances. By checking the satisfying results based on $r \wedge s \wedge t$, we can decide whether to enforce cannot-link loss $CL(x_g, x_h)$.

\section{Putting It All Together - Efficient Training Strategy}
\label{sec:training}
Our training strategy consists of two training branches and effectively has two ways of creating mini-batches for training. For instance-difficulty or global-size constraints, we treat their loss functions as addictive losses to the clustering branch so that no new branch needs to be created. For pairwise or triplet constraints, we build another output branch and train the whole network in an alternating fashion. We treat these two groups of constraints differently for a principled reason. For pairwise and triplet constraints, we have explicit constraints on instances and the composition of the clusters. This can be (and often is) contradictory (i.e., incompatible) with the clustering loss. This is indeed something we showed in our ECML 2006 paper \citep{davidson2006measuring}, where we showed that pairwise constraints could hurt clustering performance. However, since the instance level and group level constraints are guidance not explicitly on specific instances assignments, they can be folded into the clustering loss.

\textbf{Loss Branch for Instance Constraints.} In deep learning, it is common to add loss functions defined over the same output units. In the Improved DEC method \citep{guo2017improved}, the clustering loss $\ell_{C}$ and reconstruction loss $\ell_{R}$ were added together. To this, we add the instance difficulty loss $\ell_{I}$. This effectively adds guidance to speed up training convergence by identifying ``easy" instances and increase the model's robustness by ignoring ``difficult" instances. Similarly, we treat the global size constraints loss $\ell_{G}$ as an additional additive loss. All instances whether or not they are part of triplet or pairwise constraints are trained through this branch and the mini-batches are created randomly.

\textbf{Loss Branch For Complex Constraints.}
Our framework uses more complex loss functions as they define constraints on pairs and even triples of instances.
Thus we create another loss branch that contains pairwise loss $\ell_{P}$ or triplet loss $\ell_{T}$ to help the network tune the embedding which satisfies these stronger constraints. For each constraint \emph{type} we create a mini-batch consisting of only those instances having that type of constraint. For each \emph{example} of a constraint type, we feed the constrained instances through the network, calculate the loss, calculate the change in weights but do not adjust the weights. We sum the weight adjustments for all constraint examples in the mini-batch and then adjust the weights. Hence our method is an example of batch weight updating as is standard in DL for stability reasons. The whole training procedure is summarized in Algorithm \ref{alg:dccf}.

\begin{algorithm}[h]
\caption{Deep Constrained Clustering Framework}
\label{alg:dccf}
\begin{algorithmic}
\STATE {\bfseries Input:} $X$: data, $m$: maximum epochs , $k$: number of clusters, $N$: total number of batches and $N_C$: total number of constraints batches.
\STATE {\bfseries Output:} latent embeddings $Z$, cluster assignment $S$.
\smallskip
\STATE Train the stacked denosing autoencoder to obtain $Z$
\STATE Initialize centroids $\mu$ via k-means on embedding $Z$.
\FOR{$epoch=1$ {\bfseries to} $m$}
\FOR{$batch=1$ {\bfseries to} $N$}
\STATE Calculate $\ell_{C}$ via Eqn (\ref{dec_obj}), $\ell_{R}$ via Eqn (\ref{reconstruction_loss}).
\STATE Calculate $\ell_{I}$ via Eqn (\ref{instance_loss}) or $\ell_{G}$ via Eqn (\ref{global_loss}).
\STATE Calculate total loss as $\ell_{C} + \ell_{R} + \{ \ell_{I} || \ell_{G}\}$.
\STATE Update network parameters based on total loss.
\ENDFOR
\FOR{$batch=1$ {\bfseries to} $N_C$}
\STATE Calculate $\ell_{P}$ via Eqn (\ref{must_link_loss}, \ref{cannot_link_loss}) or $\ell_{T}$ via Eqn (\ref{triplet_loss}).
\STATE Update network parameters based on $\{\ell_{P} || \ell_{T}\}$ .
\ENDFOR
\STATE Forward pass to compute $Z$ and $S_i = \argmax_{j} q_{ij}$.
\ENDFOR
\end{algorithmic}
\end{algorithm}

\section{Generating Constraints from an Ontology Graph and Learning with Multiple Types of Constraints Simultaneously}
\label{sec:multiple_constraints}
In most constrained clustering works, the constraints are normally generated from ground truth labels. For example, the pairwise constraints can be generated from labeled instances by random picking each pair of points and checking the labels; the triplet constraints can be generated based on the latent embeddings' distance among the triplet (i.e., the positive instance should be close to the anchor in latent embedding space while the negative instance will be further). However, to get a good latent embedding, we still need a large amount of ground-truth labels to learn the representation. Here we seek different sources to generate constraints to add in richer side information from humans into the clustering framework to derive better clustering results.

\begin{figure}[h]
\centering
\subfigure{\includegraphics[width=1\columnwidth]{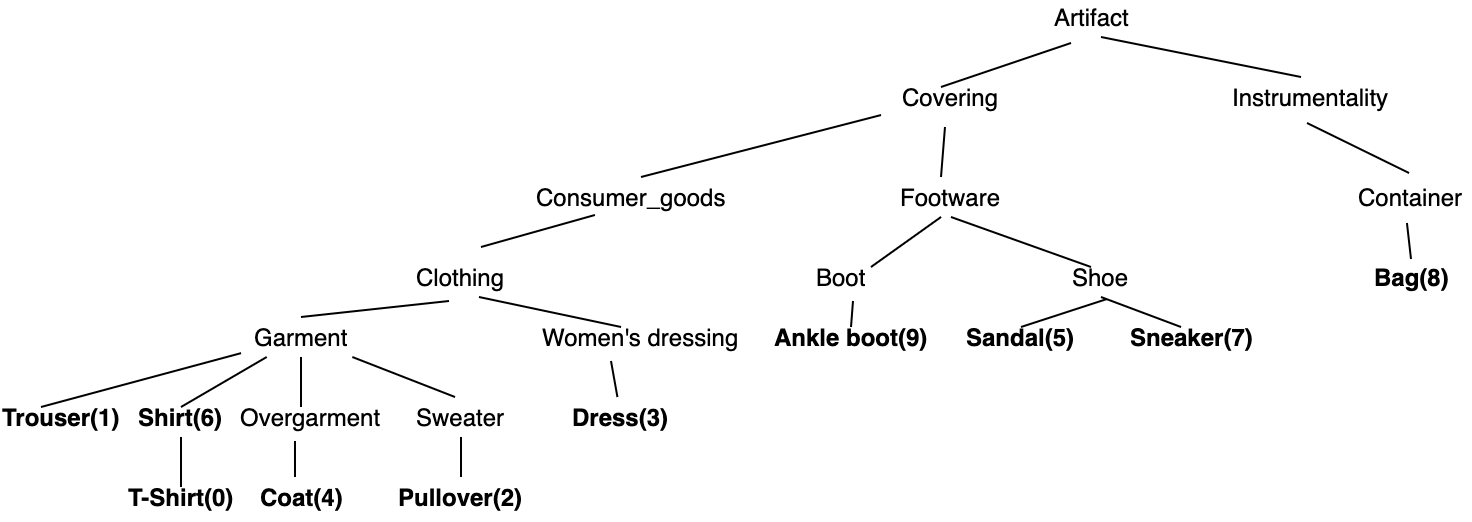}}
\caption{WordNet hierarchy of Fashion MNIST data set. The bolded classes are classes we will cluster upon. }
\label{fig:wordnet_fashion}
\end{figure}

\textbf{Generating Constraints from an Ontology Graph.}
In this section, we propose a new empirical strategy to generate triplet constraints based on ontology graphs. The class label names (i.e., \texttt{Sneaker}) can be used to place an instance in the ontology. The ontology graph (knowledge graph) represents a collection of interlinked descriptions of entities. Here we choose to use WordNet \footnote{\url{https://wordnet.princeton.edu/}}
as the ontology graph, WordNet groups English words into sets of synonyms called synsets; it also provides short definitions and usage examples and records a number of relations among these synonym sets or their members. WordNet can thus be seen as a combination of dictionary and thesaurus. We have visualized the WordNet hierarchy structure for Fashion MNIST (the data set we will use for experiments in section \ref{sec:experiments}, it consists of a training set of $60000$ examples and a test set of $10000$ examples. Each example is a $28$-by-$28$ grayscale image, associated with a label from $10$ classes. ) in Figure \ref{fig:wordnet_fashion}.

We measure the similarity between different classes based on the shortest path that connects the two classes. Given class $C_i$ and $C_j$ and the shortest path $d_{ij}$ between $C_i$ and $C_j$, the similarity is $sim(C_i, C_j) = \frac{1}{d_{ij}+1}$. Note that the similarity value ranges from $(0, 1]$, and the larger value represents a larger similarity. Given an anchor instance $a$, positive instance $p$ and negative instance $n$, we wish the class similarity between $a$ and $p$ to be as large as possible and the similarity between $a$ to $n$ to be as small as possible. To satisfy these requirements we can set the positive threshold $\theta_{p}$ to enforce the $sim(a, p) > \theta_{p}$, similarly we can set negative threshold $\theta_{n}$ to ensure both $sim(a,n) < \theta_{n}$ and $sim(p, n) < \theta_{n}$. Further we require $\theta_{n} < \theta_{p}$.
Note that when $\theta_{p}$ equals $1$, the triplet constraints will be equivalent as one Cannot-link constraint and two Must-link constraints since both the anchor and positive instance are from the same class. With proper positive and negative thresholds, we can generate triplet constraints from a set of labeled points with WordNet knowledge.

\begin{figure}[h]
\centering
\subfigure{\includegraphics[width=0.48\columnwidth]{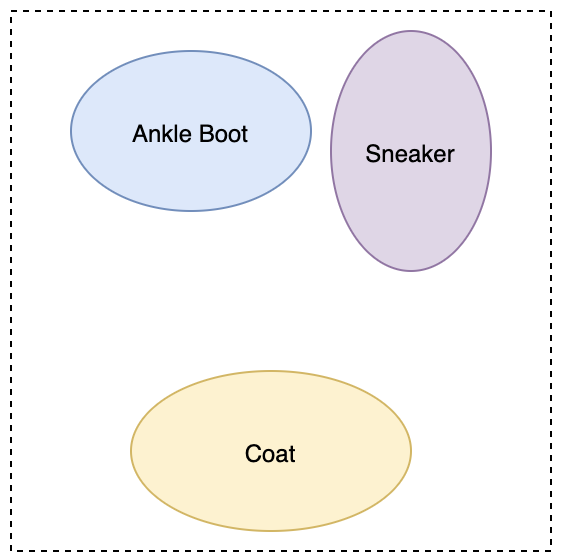}}
\hfill
\subfigure{\includegraphics[width=0.48\columnwidth]{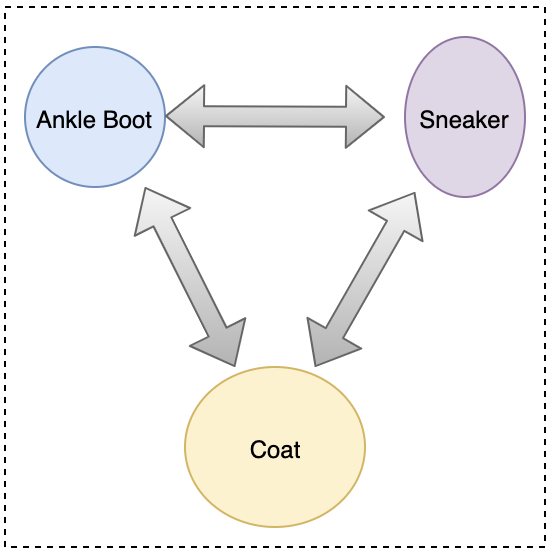}}
\caption{Examples of the ideal embedding (left-hand) learnt from an ontology using both triplets and pairwise constraints and the embedding learnt from just pairwise constraints (right-hand). Note in the later the three clusters are far apart from each other because the cannot-links (grey arrows) between these clusters will push them as far as possible which contradicts the ideal data embedding that ankle boots and sneakers are semantically similar.}
\label{fig:multiple_cons}
\end{figure}

\textbf{Learning with Multiple Types of Constraints Simultaneously.}
It is natural to wish to take advantage of all the generated constraints, even if they are different types. Here we study learning with pairwise constraints and triplet constraints together as they are most common and useful. We motivate the need to learn these two types of constraints together in Figure \ref{fig:multiple_cons}.

Given adequate pairwise constraints, the model can find correct clusters as can be seen from the right-hand side in Figure \ref{fig:multiple_cons}, but the latent semantic similarity relationships have been destroyed. Directly applying triplet constraints to this case may end up with a reasonable semantic latent space, but the circles and triangles may be overlapping without cannot-links. To learn a better embedding, we aim to add triplet constraints together with pairwise constraints to the clustering framework. In this case, we have both pairwise constraints between these three classes, as well as the triplet constraints. For example, in Figure \ref{fig:multiple_cons} the anchor and positive classes are ankle boots and sneakers whilst the negative class is coat.

Our algorithm \ref{alg:dccf} can be naturally extended to learn multiple types of constraints at the same time. Specifically, we prepare the pairwise constraints and triplet constraints with $N_p$ and $N_t$ batches in advance and then optimize the clustering loss $\ell_{C}$, pairwise loss $\ell_{P}$ and triplet loss $\ell_{T}$ in an iterative way. The entire learning process is detailed in Algorithm \ref{alg:pairwise_triplet}.
\begin{algorithm}[h]
\caption{Learning with Pairwise and Triplet Constraints}
\label{alg:pairwise_triplet}
\begin{algorithmic}
\STATE {\bfseries Input:} $X$: data, $m$: maximum epochs , $k$: number of clusters, $N$: total number of batches and $N_p$: total number of pairwise constraints batches, $N_t$: total number of triplet constraints batches.
\STATE {\bfseries Output:} latent embeddings $Z$, cluster assignment $S$.
\smallskip
\STATE Train the stacked denosing autoencoder to obtain $Z$
\STATE Initialize centroids $\mu$ via k-means on embedding $Z$.
\FOR{$epoch=1$ {\bfseries to} $m$}
\FOR{$batch=1$ {\bfseries to} $N$}
\STATE Calculate $\ell_{C}$ via Eqn (\ref{dec_obj}), $\ell_{R}$ via Eqn (\ref{reconstruction_loss}).
\STATE Calculate total loss as $\ell_{C} + \ell_{R}$.
\STATE Update network parameters based on total loss.
\ENDFOR
\FOR{$batch=1$ {\bfseries to} $N_p$}
\STATE Calculate $\ell_{P}$ via Eqn (\ref{must_link_loss}, \ref{cannot_link_loss}).
\STATE Update network parameters based on $\ell_{P}$.
\ENDFOR
\FOR{$batch=1$ {\bfseries to} $N_t$}
\STATE Calculate $\ell_{T}$ via Eqn (\ref{triplet_loss}).
\STATE Update network parameters based on $ \ell_{T}$ .
\ENDFOR
\STATE Forward pass to compute $Z$ and $S_i = \argmax_{j} q_{ij}$.
\ENDFOR
\end{algorithmic}
\end{algorithm}

\section{Experiments}
\label{sec:experiments}
All data and code used to perform these experiments are available online (\url{http://github.com/blueocean92/deep_constrained_clustering}) to help with reproducibility. In our experiments, we aim to address the following questions:
\begin{itemize}
\item How does our end-to-end deep clustering approach using traditional pairwise constraints compare with traditional constrained clustering methods? The latter is given the same auto-encoding representation $Z$ used to initialize our method. (see Table \ref{tab:pairwise_neg})
\item Are the new types of constraints we create for the deep clustering method useful in practice? (see Section \ref{exp:instance}, \ref{exp:triplet}, \ref{exp:global})
\item Is our end-to-end deep constrained clustering method more robust to the well known negative effects of constraints we published earlier \citep{davidson2006measuring}? How our learned embedding overcomes the negative effects of constraints? (see Section \ref{exp:pairwise})
\item How the model performs with constraints generated from ontologies? (see Section \ref{exp:pairwise_triplet})
\item How is the proposed model's robustness towards noisy constraints? (see Section \ref{exp:robustness})
\item How do the different components of our approach contribute to our final performance? (see our Ablation study in Section \ref{exp:ablation_study})
\item How is the scalability of our proposed framework? (see Section \ref{exp:runtime_study})

\end{itemize}

\subsection{Datasets}
To study the performance and generality of different algorithms, we evaluate the proposed method on two image datasets and one test dataset:

\noindent
\textbf{MNIST}: Consists of $70000$ handwritten digits of $28$-by-$28$ pixel size. The digits are centered and size-normalized in our experiments \citep{lecun1998gradient}.

\noindent
\textbf{FASHION-MNIST}: A Zalando's article images-consisting of a training set of $60000$ examples and a test set of $10000$ examples. Each example is a $28$-by-$28$ grayscale image, associated with a label from $10$ classes.

\noindent
\textbf{REUTERS-10K}: This dataset contains English news stories labeled with a category tree \citep{lewis2004rcv1}. To be comparable with the previous baselines, we used $4$ root categories: \texttt{corporate/industrial}, \texttt{government/social}, \texttt{markets} and \texttt{economics} as labels and excluded all documents with multiple labels. We randomly sampled a subset of $10000$ examples and computed TF-IDF features on the $2000$ most common words. 

\subsection{Evaluation Metric}
We adopt standard metrics for evaluating clustering performance which measure how close the clustering found is to the ground truth result. Specifically, we employ the following two metrics: normalized mutual information (\textbf{NMI}) \citep{strehl2000impact,xu2003document} and clustering accuracy (\textbf{Acc}) \citep{xu2003document}. For data point $x_i$, let $l_i$ and $c_i$ denote its true label and predicted cluster respectively. Let $l = (l_1, ... l_n)$ and similarity $c = (c_1, ... c_n)$. \textbf{NMI} is defined as:
\begin{equation*}
\label{NMI}
\textbf{NMI}(l, c) = \frac{\textbf{MI}(l, c)}{max\{H(l), H(c)\}}
\end{equation*}
where $\textbf{MI}(l, c)$ denotes the mutual information between $l$ and $c$, and $H$ denotes their entropy. The \textbf{Acc} is defined as:
\begin{equation*}
\label{acc}
\textbf{Acc}(l, c) = \max_{m} \frac{\sum_{i=1}^n \mathbf{1}\{l_i = m(c_i)\}}{n}
\end{equation*}
where $m$ ranges over all possible one-to-one mappings between clusters and labels. The optimal assignment of $m$ can be computed using the Kuhn-Munkres algorithm \citep{munkres1957algorithms}. Both metrics are commonly used in the clustering literature and with higher values indicating better clustering results. By using them together we get a better understanding of the effectiveness of the clustering algorithms.

\subsection{Implementation Details}
\textbf{Basic Deep Clustering Implementation.} To be comparable with deep clustering baselines, we set the encoder network as a fully connected multilayer perceptron with dimensions $d-500-500-2000-10$ for all datasets, where $d$ is the dimension of input data(features). The decoder network is a mirror of the encoder. All the internal layers are activated by the ReLU \citep{nair2010rectified} nonlinearity function. For a fair comparison with baseline methods, we used the same greedy layer-wise pre-training strategy to calculate the auto-encoders embedding. To initialize clustering centroids, we run k-means with 20 restarts and select the best solution. We choose Adam optimizer with an initial learning rate of $0.001$ for all the experiments. We adopt standard metrics for evaluating clustering performance, which measures how close the clustering found is to the ground truth result. Specifically, we employ the following two metrics: normalized mutual information (\textbf{NMI}) \citep{strehl2000impact,xu2003document} and clustering accuracy (\textbf{Acc}) \citep{xu2003document}. In our baseline comparisons, we use IDEC \citep{guo2017improved}, a non-constrained improved version of DEC published recently.

\textbf{Pairwise Constraints Experiments.} We randomly select pairs of instances and generate the corresponding pairwise constraints between them.
To ensure transitivity, we calculate the transitive closure over all must-linked instances and then generate entailed constraints from the cannot-link constraints \citep{davidson2007intractability}. Since our loss function for must-link constraints is combined with reconstruction loss, we use grid search and set the penalty weight for must-link as $0.1$.

\begin{figure}[h]
\centering
\subfigure{\includegraphics[width=0.6\columnwidth]{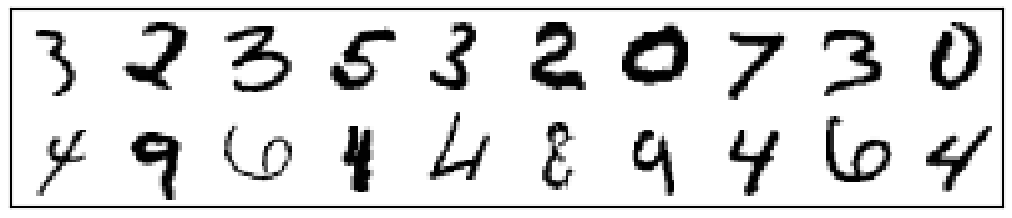}}
\caption{Example of instance difficulty constraints. Top row shows the ``easy'' instances and second row shows the ``difficult'' instances. }
\label{fig:instance_mnist}
\end{figure}

\textbf{Instance Difficulty Constraints Experiments.} To simulate human-guided instance difficulty constraints, we use k-means as a weak base learner and mark all the incorrectly clustered instances as difficult with confidence $0.1$, we also mark the correctly classified instances as accessible instances with confidence $1$.
In Figure \ref{fig:instance_mnist}, we give some example difficulty constraints found using this method.

\begin{figure}[h]
\centering
\subfigure{\includegraphics[width=0.48\columnwidth]{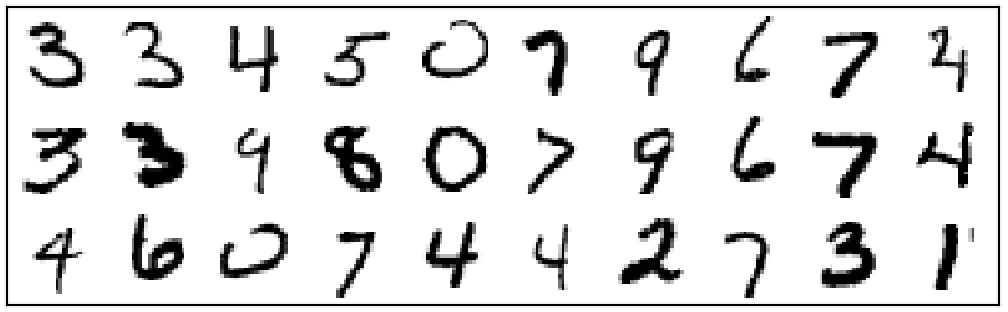}}
\hfill
\subfigure{\includegraphics[width=0.48\columnwidth]{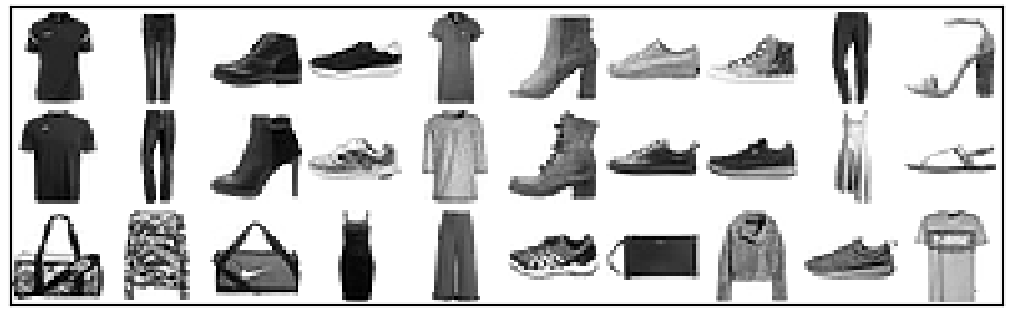}}
\caption{Examples of the generated triplet constraints for MNIST and Fashion. The three rows for each plot shows the anchor instances, positive instances and negative instances correspondingly.}
\label{fig:triplet_visual}
\end{figure}

\textbf{Triplet Constraints Experiments.} Triplet constraints can state that instance $i$ is more similar to instance $j$ than instance $k$. To simulate human guidance on triplet constraints, we randomly select $n$ instances as anchors ($i$); for each anchor, we randomly select two instances ($j$ and $k$) based on the similarity between the anchor. The similarity is calculated as the euclidian distance $d$ between two instances pre-trained embedding. The pre-trained embedding is extracted from our deep clustering network trained with $100000$ pairwise constraints.
Figure \ref{fig:triplet_visual} shows the generated triplets constraints. Through grid search we set the triplet loss margin $\theta = 0.1$.

\textbf{Global Size Constraints Experiments.} We apply global size constraints to MNIST and Fashion datasets since they satisfy the balanced size assumptions. The total number of clusters is set to $10$, and each class has the same number of instances.

\subsection{Experimental Results}
\label{sec:exp_results}
\subsubsection{Experiments on instance difficulty.}
\label{exp:instance}
\begin{table}[htb]
\caption{Left table shows baseline results for Improved DEC \citep{guo2017improved} averaged over $20$ trials. Right table lists experiments using instance difficulty constraints (mean $\pm$ std) averaged over $20$ trials. }
\begin{minipage}{.48\linewidth}
\centering
\resizebox{\columnwidth}{!}{%
\begin{tabular}{cccc}
\toprule
& MNIST & Fashion & Reuters\\
\midrule
Acc(\%) & $88.29 \pm 0.05$ &$58.74 \pm 0.08$ &$75.20 \pm 0.07$\\
NMI(\%) & $86.12 \pm 0.09$ &$63.27 \pm 0.11$ &$54.16 \pm 1.73$ \\
Epoch & $87.60 \pm 12.53$ & $77.20 \pm 11.28$ &$12.90 \pm 2.03$\\
\bottomrule
\end{tabular}
}
\end{minipage}
\hfill
\begin{minipage}{.48\linewidth}
\centering
\resizebox{\columnwidth}{!}{%
\begin{tabular}{cccc}
\toprule
& MNIST & Fashion & Reuters\\
\midrule
Acc(\%) & $91.02 \pm 0.34$ &$62.17 \pm 0.06$ &$78.01 \pm 0.13$\\
NMI(\%) & $88.08 \pm 0.14$ &$64.95 \pm 0.04$ &$56.02 \pm 0.21$ \\
Epoch & $29.70 \pm 4.25$ & $47.60 \pm 6.98$ &$9.50 \pm 1.80$\\
\bottomrule
\end{tabular}
}
\end{minipage}
\label{tab:instance}
\end{table}

In Table \ref{tab:instance}, we report the average test performance of the deep clustering framework without any constraints on the left. In comparison, we report the average test performance of deep clustering framework with instance difficulty constraints on the right, and we find the model learned with instance difficulty constraints outperforms the baseline method in all datasets. This is to be expected as we have given the algorithm more information than the baseline method, but it demonstrates our method can make good use of this extra information. What is unexpected is the effectiveness of speeding up the learning process and will be the focus of future work.

\begin{figure*}[th]
\centering
\subfigure[MNIST]{\includegraphics[width=0.325\textwidth]{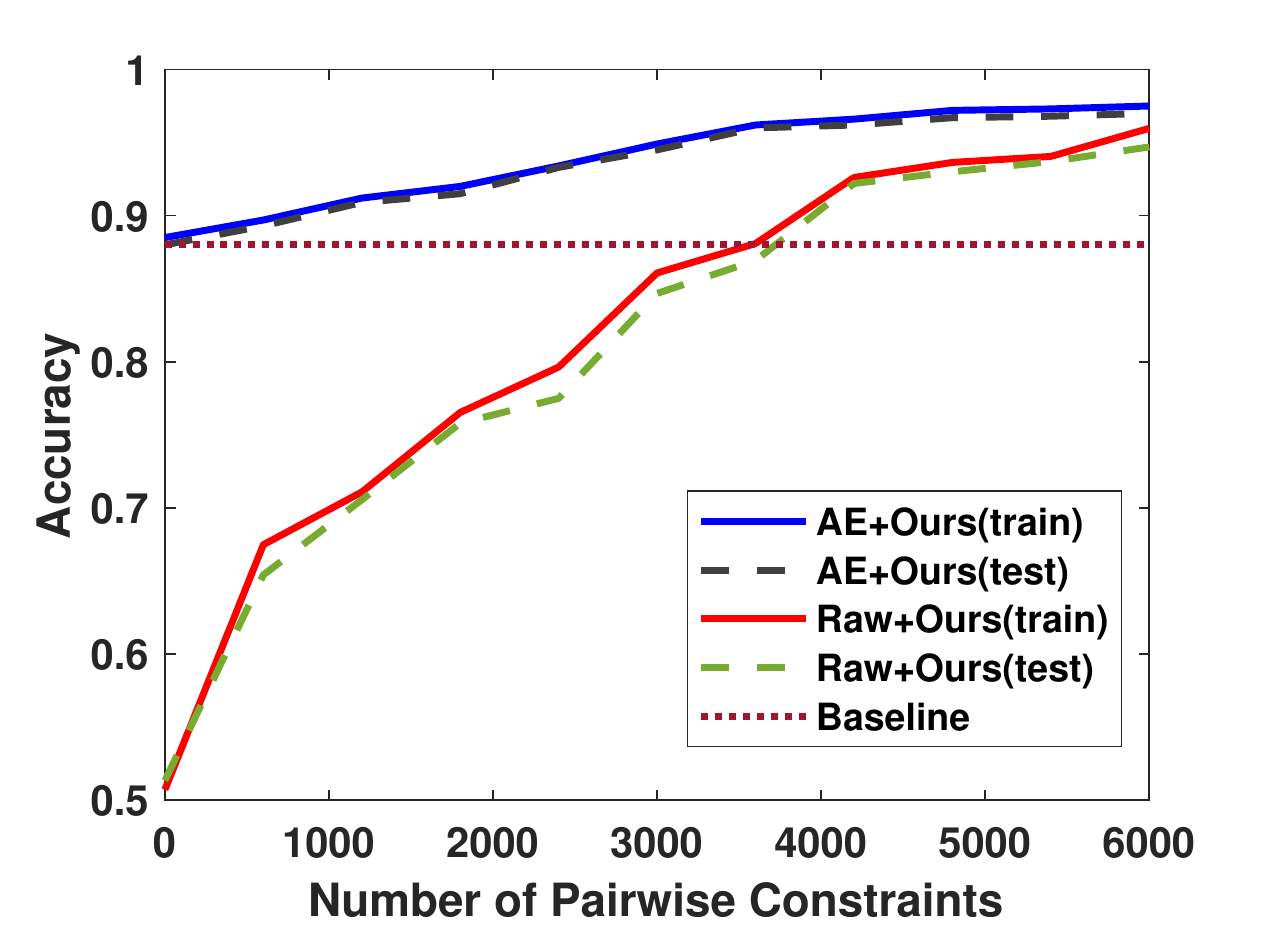}}
\hfill
\subfigure[Fashion]{\includegraphics[width=0.325\textwidth]{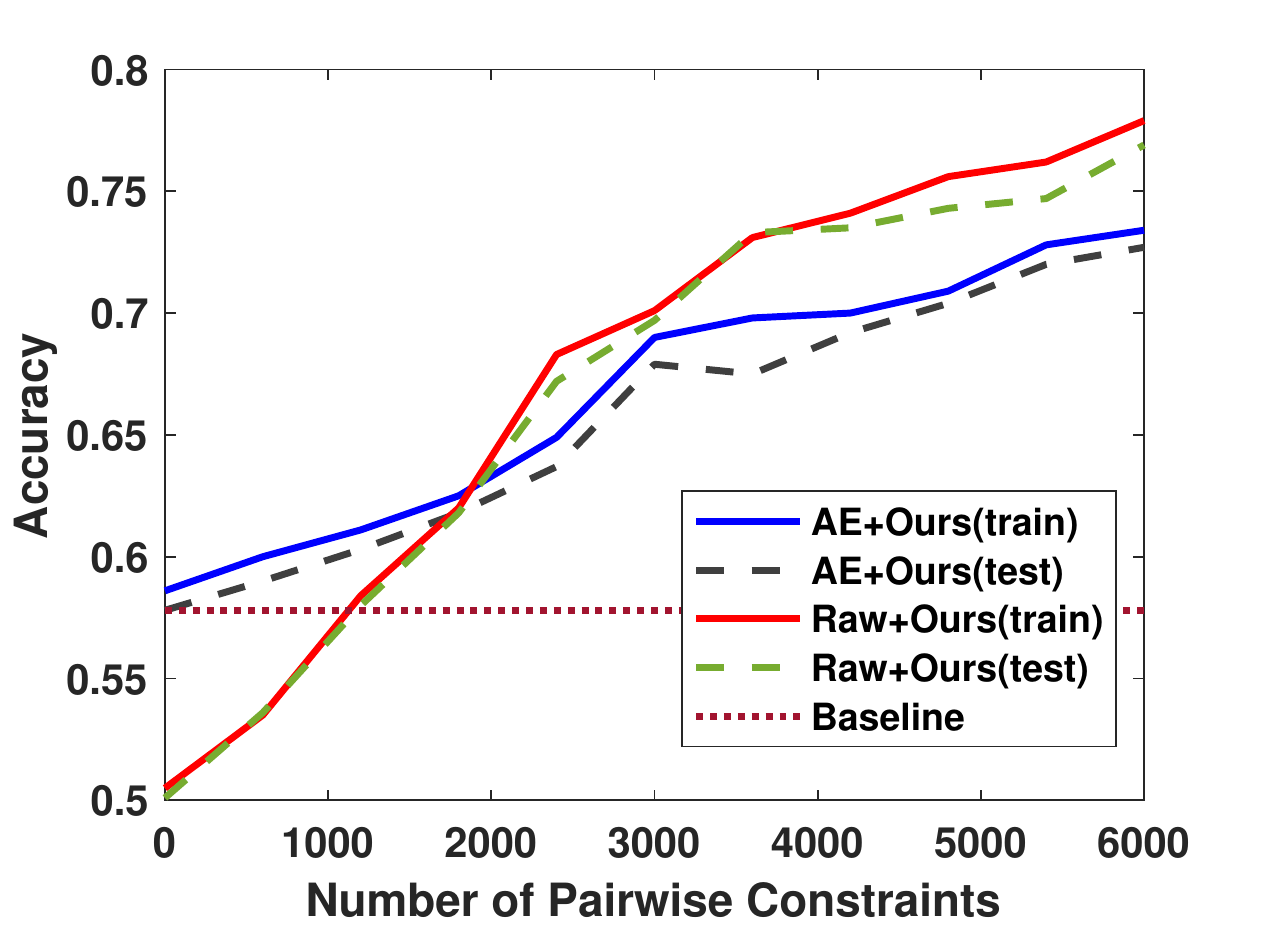}}
\hfill
\subfigure[Reuters]{\includegraphics[width=0.325\textwidth]{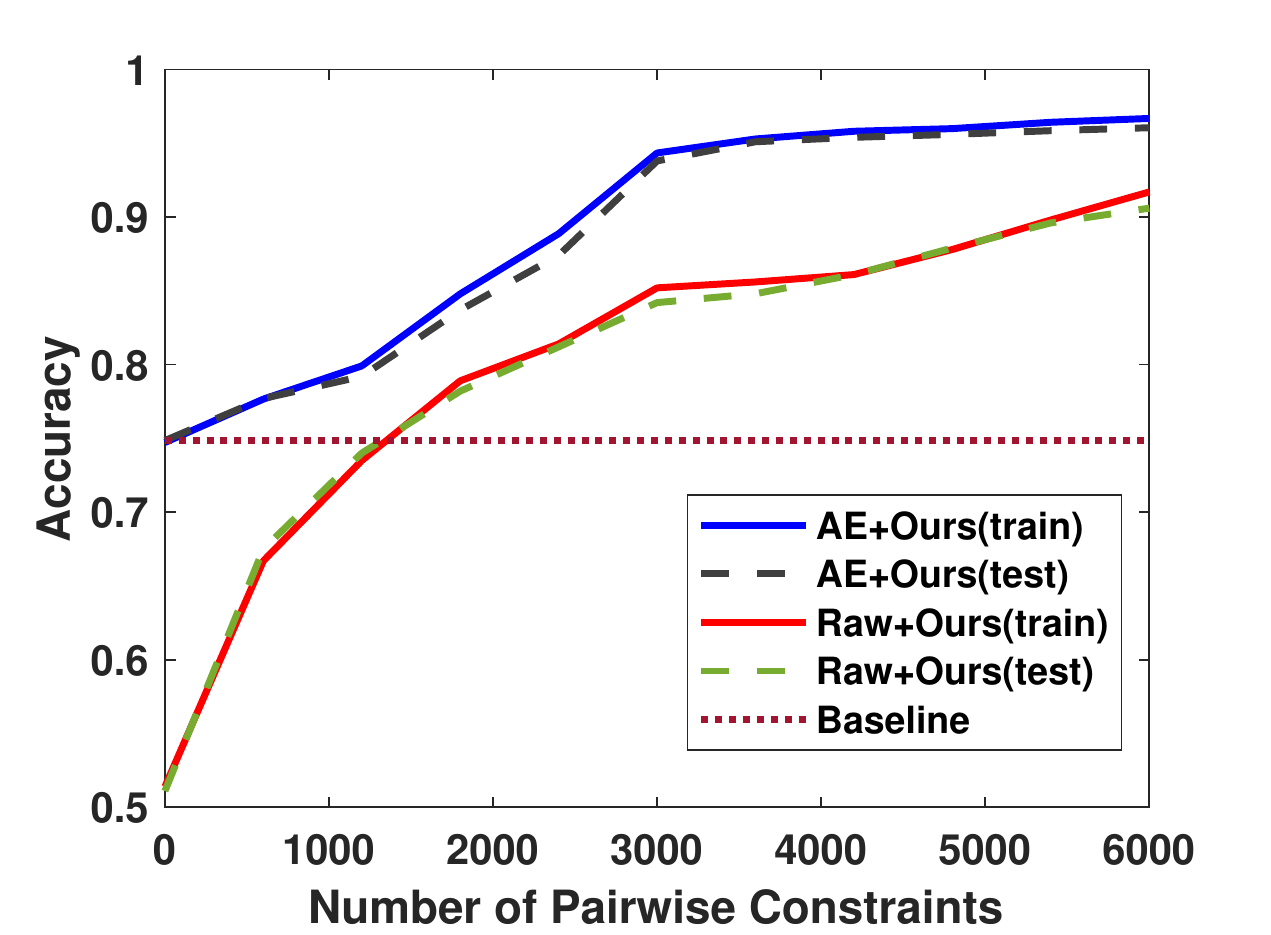}}

\subfigure[MNIST]{\includegraphics[width=0.325\textwidth]{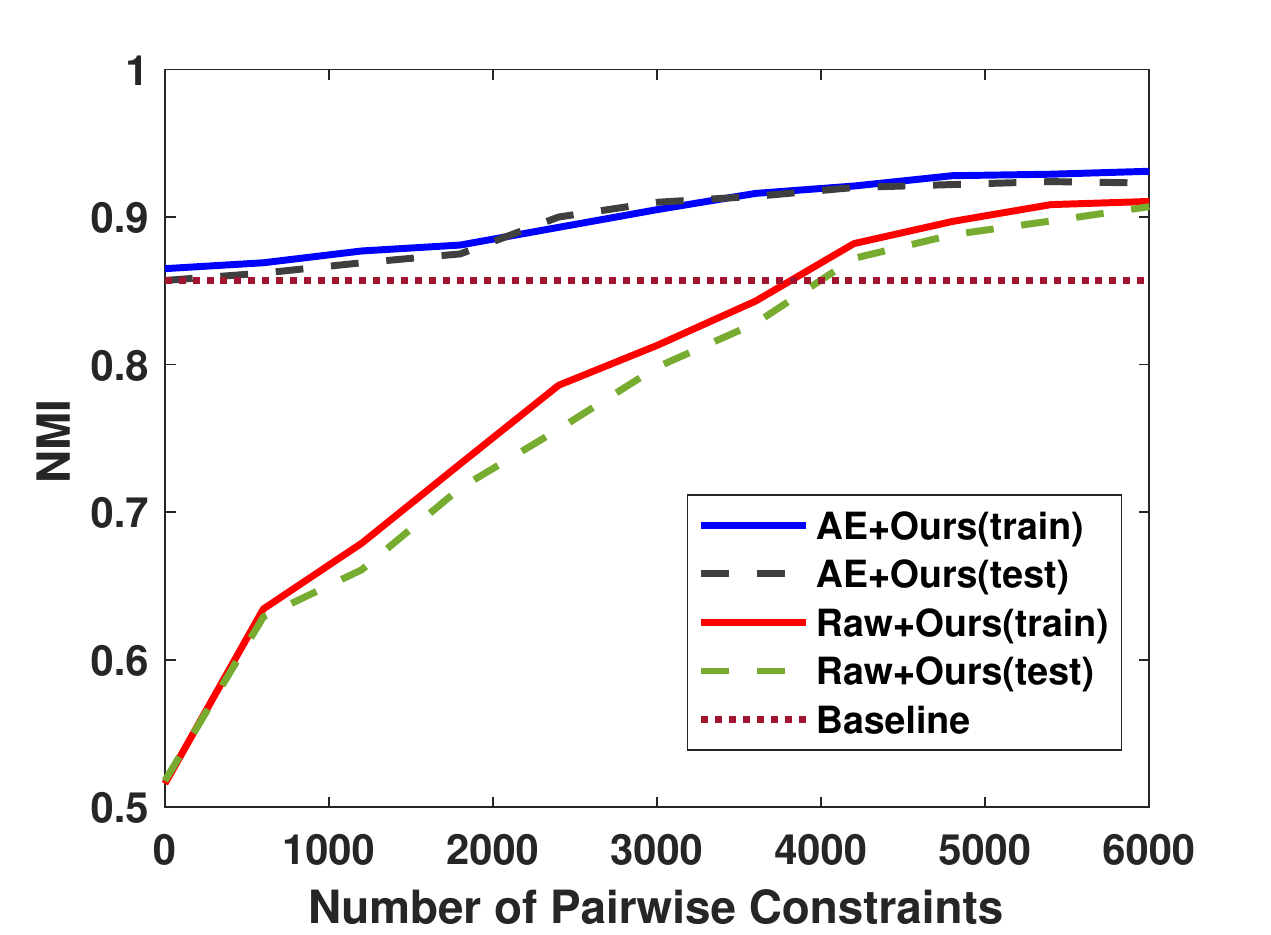}}
\hfill
\subfigure[Fashion]{\includegraphics[width=0.325\textwidth]{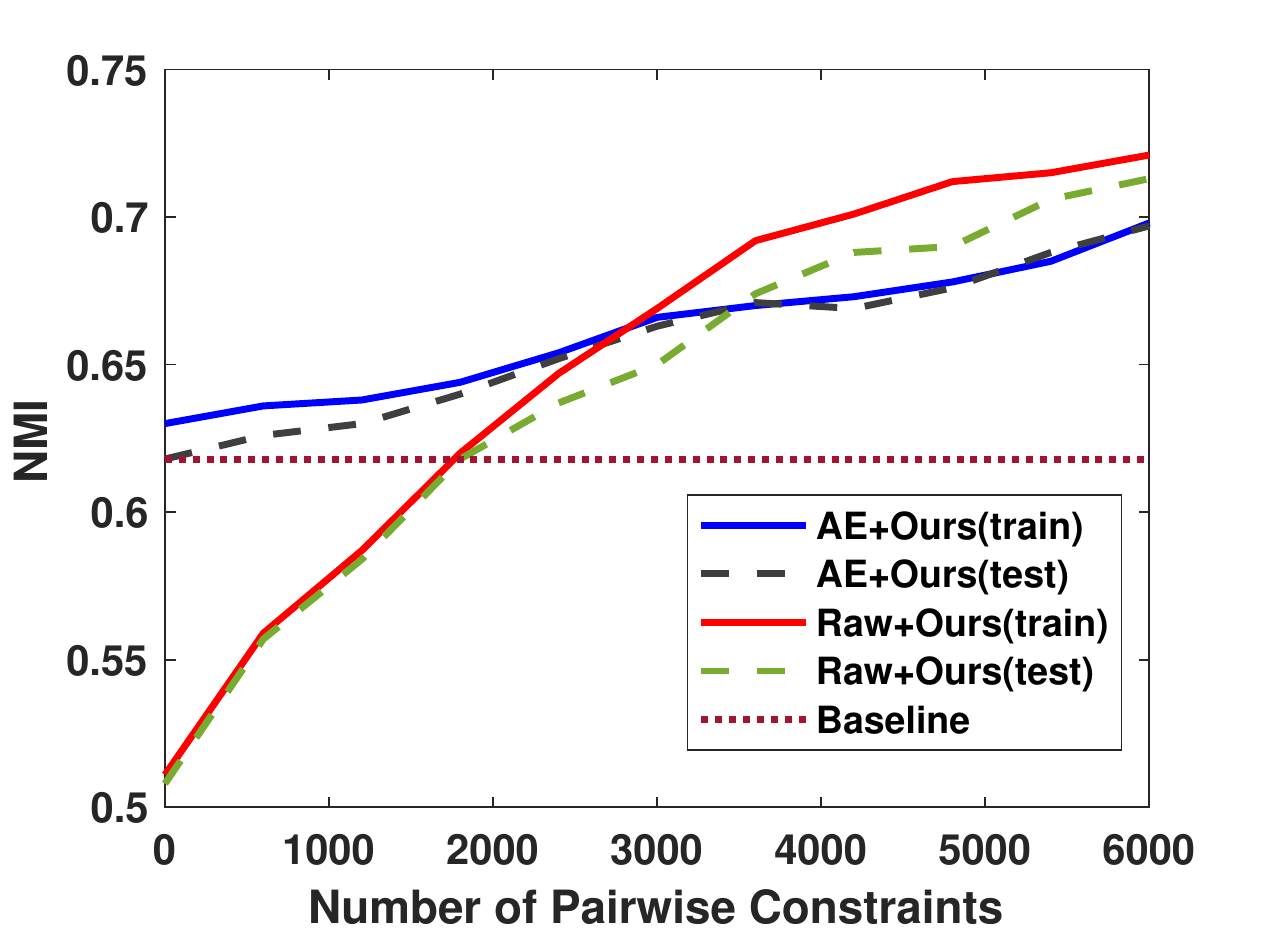}}
\hfill
\subfigure[Reuters]{\includegraphics[width=0.325\textwidth]{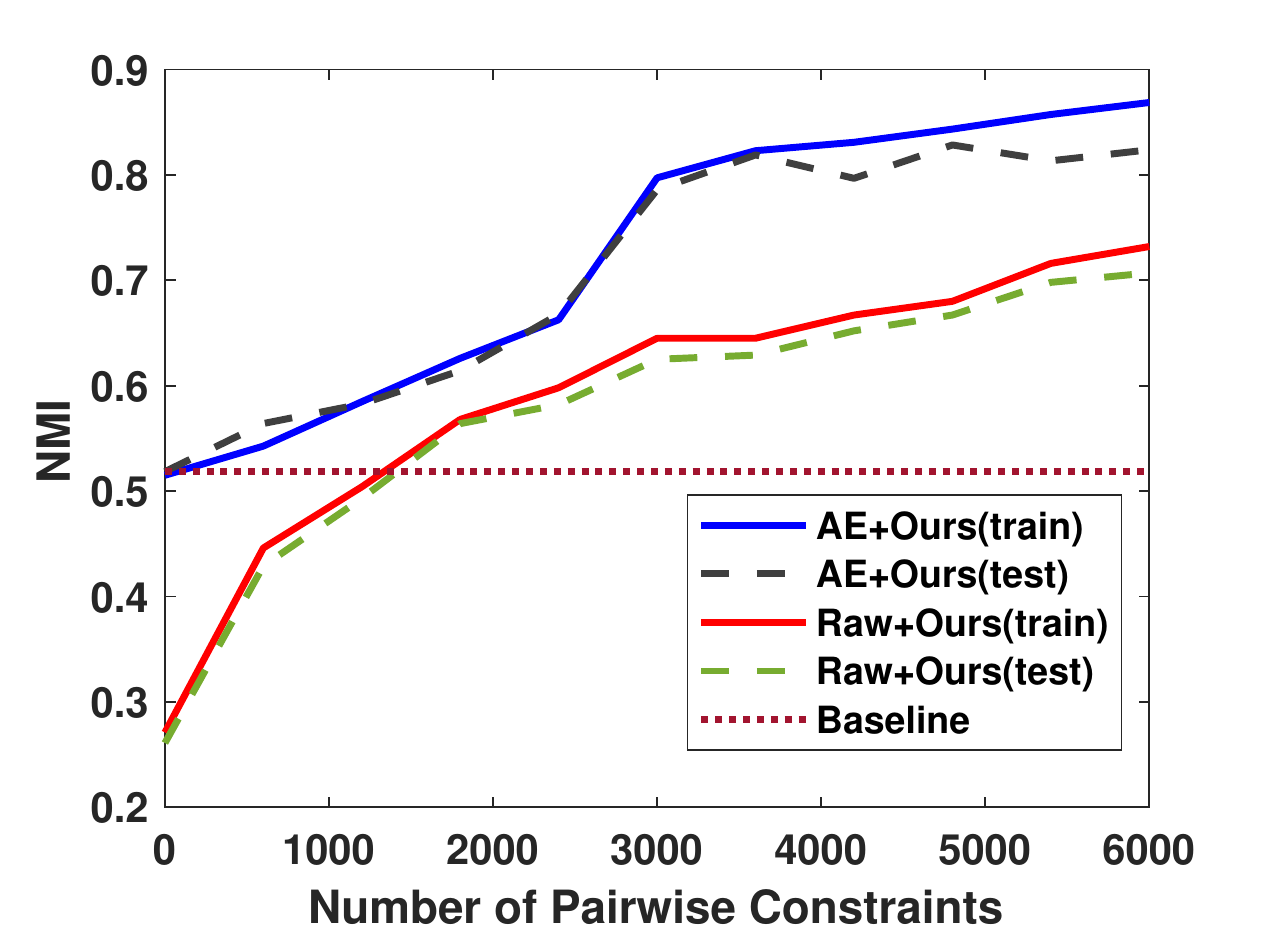}}
\caption{Clustering accuracy and NMI on training test sets for different number of pairwise constraints. AE means an autoencoder was used to seed our method. The horizontal maroon colored baseline shows the IDEC's \citep{guo2017improved} test set performance.}
\label{fig:pairwise_performance}
\end{figure*}

\subsubsection{Experiments on pairwise constraints}
\label{exp:pairwise}

We randomly generate $6000$ pairs of constraints which are small fractions of possible pairwise constraints for MNIST ($0.0002 \%$), Fashion ($0.0002 \%$), and Reuters ($0.006 \%$).
Recall the DEC method is initialized with auto-encoder features. To better understand the contribution of pairwise constraints, we have tested our method with both auto-encoders features and raw data. As can be seen from Figure \ref{fig:pairwise_performance}: the clustering performance improves consistently as the number of constraints increases in both settings. Moreover, with just $6000$ pairwise constraints, the performance on Reuters and MNIST increased significantly especially for the setup with raw data. We also notice that learning with raw data in Fashion achieves a better result than using autoencoder's features. This shows that the autoencoder's features may not always be suitable for DEC's clustering objective. Overall our results show pairwise constraints can help reshape the representation and improve the clustering results.

We also compare the results with recent work \citep{hsu2015neural}: our approach(autoencoders features) outperforms the best clustering accuracy reported for MNIST by a margin of $16.08\%$, $2.16\%$ and $0.13\%$ respectively for 6, 60, and 600 samples/class. Unfortunately, we can't make a comparison with Fogel's algorithm \citep{fogel2019clustering} due to an issue in their code repository.

\begin{table}[ht]
\begin{center}
\caption{Pairwise constrained clustering performance (mean $\pm$ std) averaged over $100$ constraints sets. Due to the scalability issues we apply flexible CSP with downsampled data($3000$ instances and $180$ constraints). Negative ratio is the fraction of times using constraints resulted in poorer results than not using constraints. See Figure \ref{fig:visual_embedding} and text for an explanation why our method performs well.}
\begin{tabular}{ccccc}
\toprule[1.6pt]
&Flexible CSP{*}& COP-KMeans & MPCKMeans & Ours \\
\midrule
MNIST Acc & $0.628 \pm 0.07$& $0.816 \pm 0.06$ & $0.846 \pm 0.04$ &{$\textbf{0.963} \pm \textbf{0.01}$} \\
MNIST NMI & $0.587 \pm 0.06$& $0.773 \pm 0.02$ & $0.808 \pm 0.04$ &{$\textbf{0.918} \pm \textbf{0.01}$} \\
Negative Ratio & $19 \%$& $45 \%$ & $11 \%$ &{$\textbf{0 \%}$}\\
\midrule
Fashion Acc &$0.417 \pm 0.05$ & $0.548 \pm 0.04$ & $0.589 \pm 0.05$ &{$\textbf{0.681} \pm \textbf{0.03}$}\\
Fashion NMI &$0.462 \pm 0.03$ & $0.589 \pm 0.02$ & $0.613 \pm 0.04$ &{$\textbf{0.667} \pm \textbf{0.02}$}\\
Negative Ratio & $23 \%$& $27 \%$ & $37 \%$ &{$\textbf{6 \%}$}\\
\midrule
Reuters Acc &$0.554 \pm 0.07$ & $0.712 \pm 0.04$ & $0.763 \pm 0.05$ &{$\textbf{0.950} \pm \textbf{0.02}$}\\
Reuters NMI &$0.410 \pm 0.05$ & $0.478 \pm 0.03$ & $0.544 \pm 0.04$ &{$\textbf{0.815} \pm \textbf{0.02}$}\\
Negative Ratio &$28 \%$ & $73 \%$ & $80 \%$ &{$\textbf{0 \%}$}\\
\bottomrule[1.6pt]
\label{tab:pairwise_neg}
\end{tabular}
\end{center}
\end{table}

\textbf{Negative Effects of Constraints.} Our earlier work \citep{davidson2006measuring} showed that for traditional constrained clustering algorithms, that the addition of constraints \emph{on average} helps clustering but many individual constraint sets can hurt performance in that performance is worse than using \textbf{no} constraints. Here we recreate these results even when these classic methods use auto-encoded representations. In Table \ref{tab:pairwise_neg}, we report the average performance with $3600$ randomly generated pairwise constraints. For each dataset, we randomly generated $100$ sets of constraints to test the negative effects of constraints \citep{davidson2006measuring}. In each run, we fixed the random seed and the initial centroids for k-means based methods. For each method, we compare its performance between the constrained version to the unconstrained version. We calculate the negative ratio, which is the fraction of times that the unconstrained version produced better results than the constrained version. As can be seen from the table, our proposed method achieves significant improvements than traditional non-deep constrained clustering algorithms \citep{wagstaff2001constrained,bilenko2004integrating,wang2010flexible}.

\begin{figure}[th]
\centering
\subfigure[MNIST (AE)]{\includegraphics[width=0.325\columnwidth]{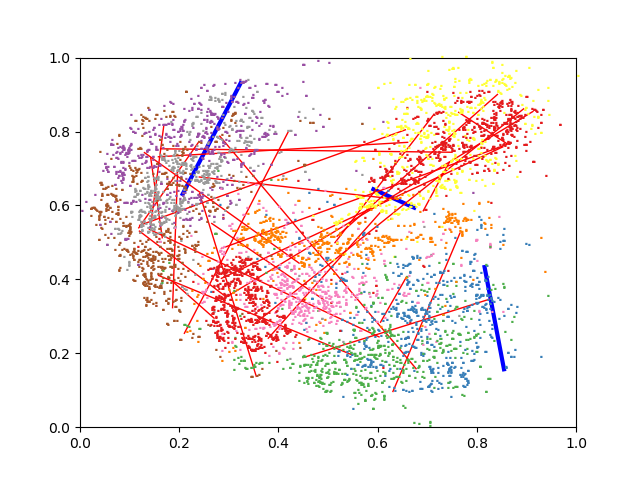}}
\hfill
\subfigure[MNIST (IDEC)]{\includegraphics[width=0.325\columnwidth]{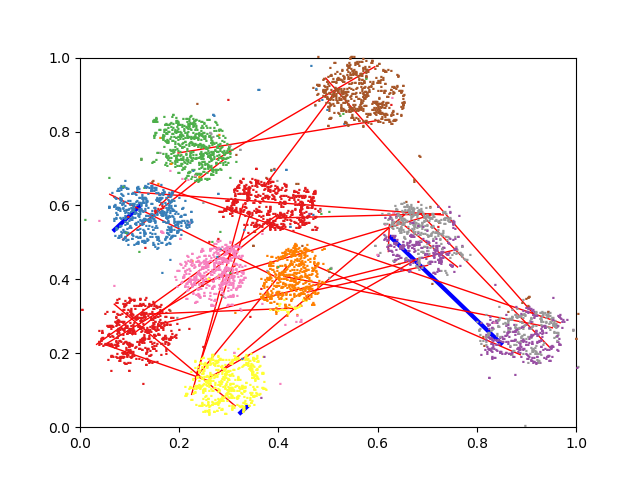}}
\hfill
\subfigure[MNIST (Ours)]{\includegraphics[width=0.325\columnwidth]{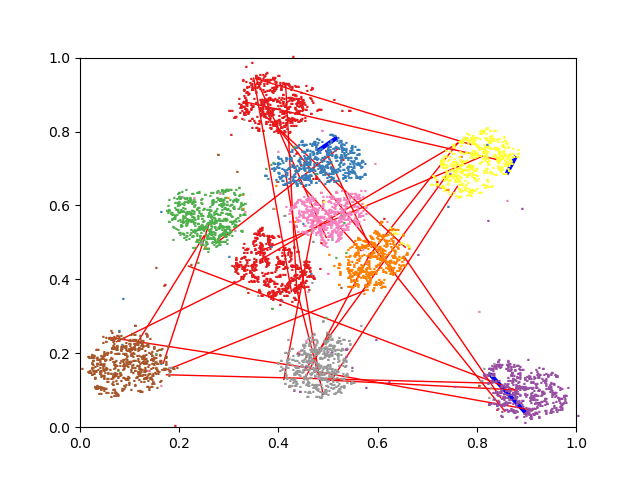}}

\subfigure[Fashion (AE)]{\includegraphics[width=0.325\columnwidth]{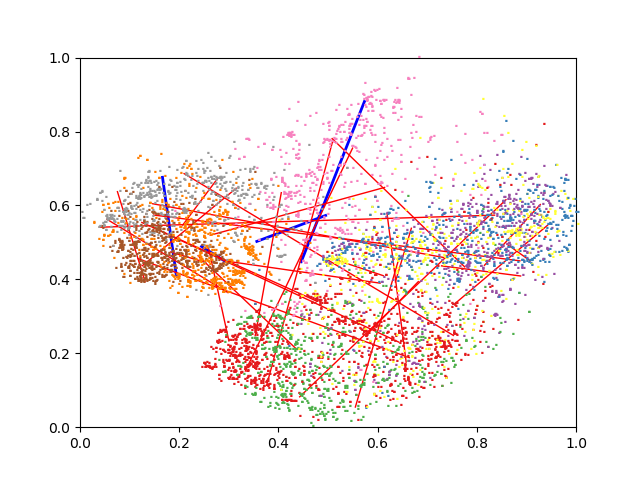}}
\hfill
\subfigure[Fashion (IDEC)]{\includegraphics[width=0.325\columnwidth]{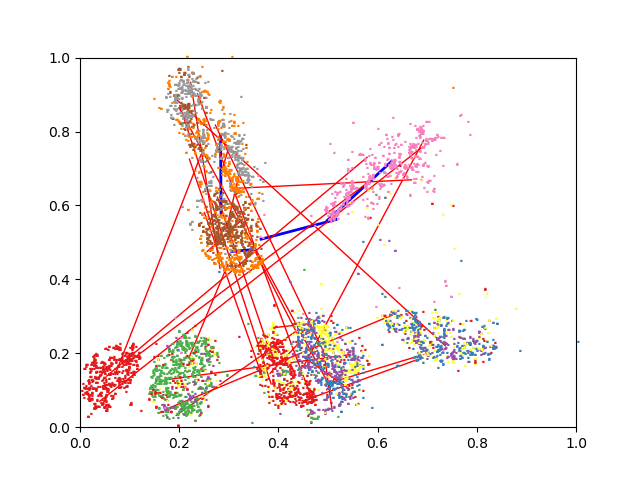}}
\hfill
\subfigure[Fashion (Ours)]{\includegraphics[width=0.325\columnwidth]{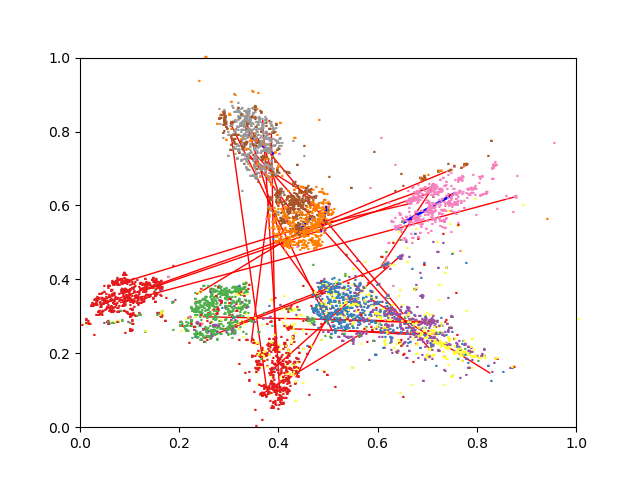}}

\subfigure[Reuters (AE)]{\includegraphics[width=0.325\columnwidth]{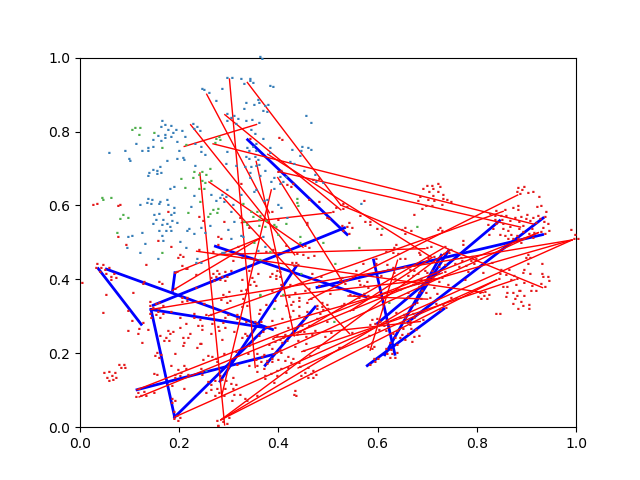}}
\hfill
\subfigure[Reuters (IDEC)]{\includegraphics[width=0.325\columnwidth]{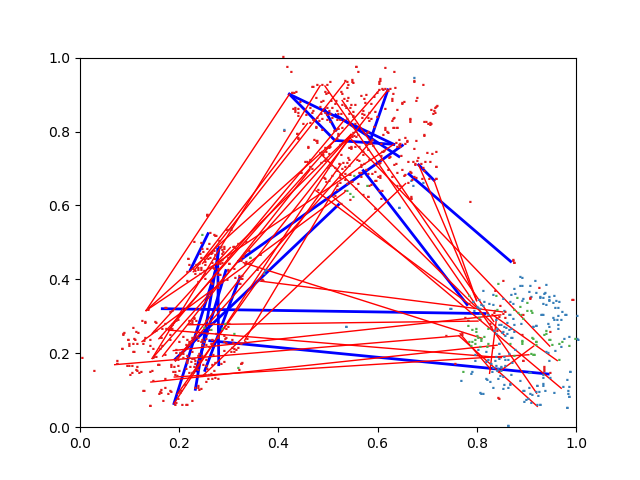}}
\hfill
\subfigure[Reuters (Ours)]{\includegraphics[width=0.325\columnwidth]{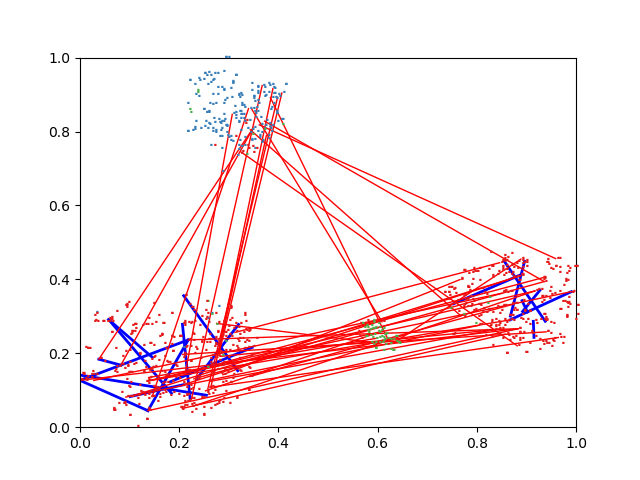}}
\caption{We visualize (using t-SNE) the latent representation for a subset of instances and pairwise constraints, we visualize the same instances and constraints for each row. The red lines are cannot-links and blue lines are must-links.}
\label{fig:visual_embedding}
\end{figure}

\begin{figure}[th]
\centering
{\includegraphics[width=0.99\columnwidth]{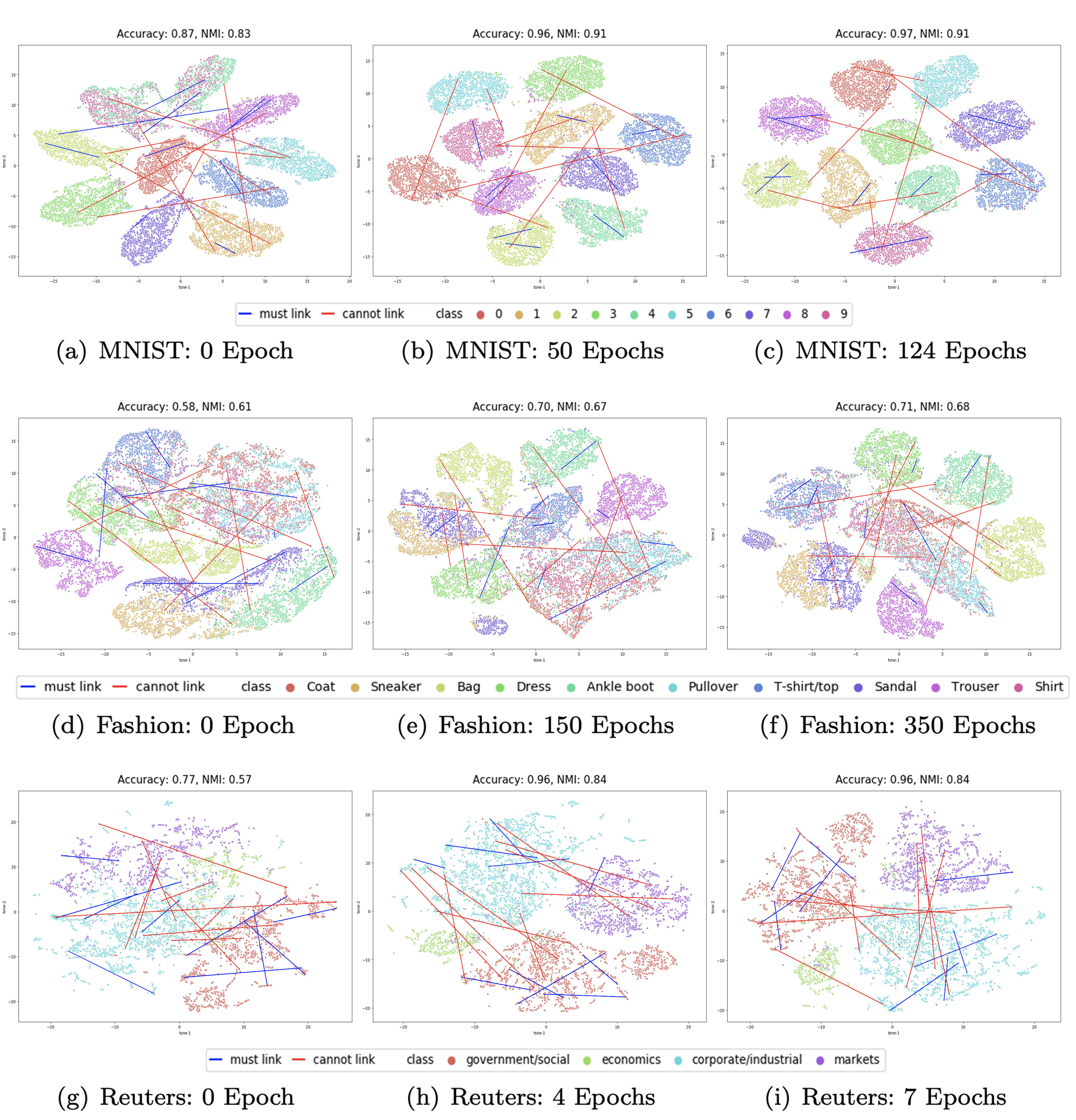}}

\caption{The embedding for a subset of instances and inconsistent pairwise constraints after several training epochs}
\label{fig:embedding_change}
\end{figure}

To understand why our method was robust to variations in constraint sets, we visualized the embeddings learned. Figure \ref{fig:visual_embedding} shows the embedded representation of a random subset of instances and its corresponding pairwise constraints using t-SNE and the learned embedding $z$. Based on Figure \ref{fig:visual_embedding}, we can see the autoencoders embedding is noisy, and lot's of constraints are inconsistent based on our earlier definition \citep{davidson2006measuring}. Further, we visualize the IDEC's latent embedding and find out the clusters are better separated. However, the inconsistent constraints still exist (blue lines across different clusters and redlines within a cluster); these constraints tend to have negative effects on traditional constrained clustering methods. Finally, for our method's results we can see the clusters are well separated, the must-links are well satisfied (blue lines are within the same cluster), and cannot-links are well satisfied (red lines are across different clusters). Hence we can conclude that end-to-end-learning can address these negative effects of constraints by simultaneously learning a representation that is consistent with the constraints and clustering the data.
This result has profound practical significance as practitioners typically only have one constraint set to work with.

To fully understand how our constrained clustering model finds a new representation to satisfy those constraints, we have visualized the latent embeddings during the training process in Figure \ref{fig:embedding_change}.

\subsubsection{Experiments on triplet constraints}
\label{exp:triplet}
\begin{figure}[h]
\centering
\subfigure[MNIST(Triplet)]{\includegraphics[width=0.45\columnwidth]{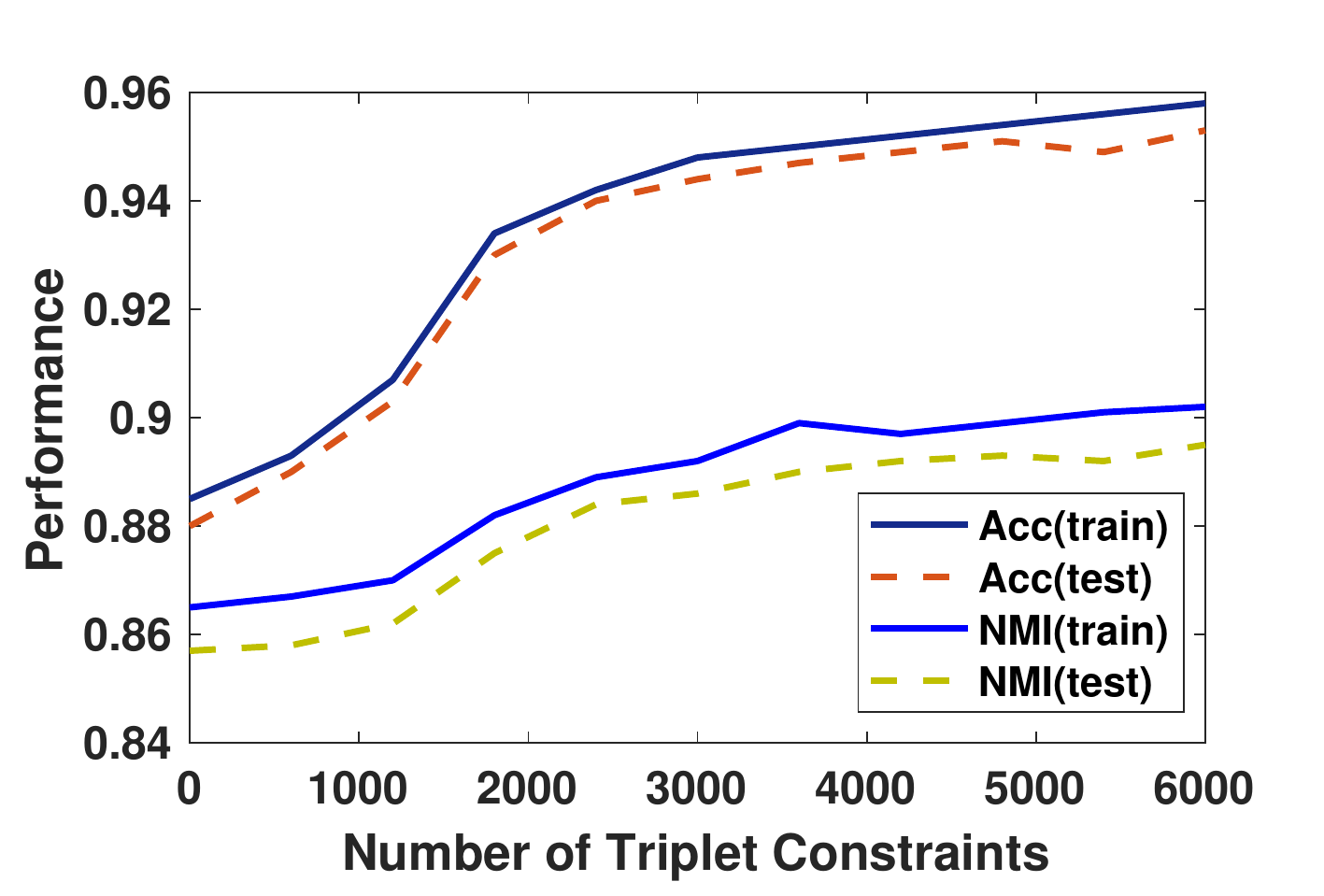}}
\hfill
\subfigure[Fashion(Triplet)]{\includegraphics[width=0.45\columnwidth]{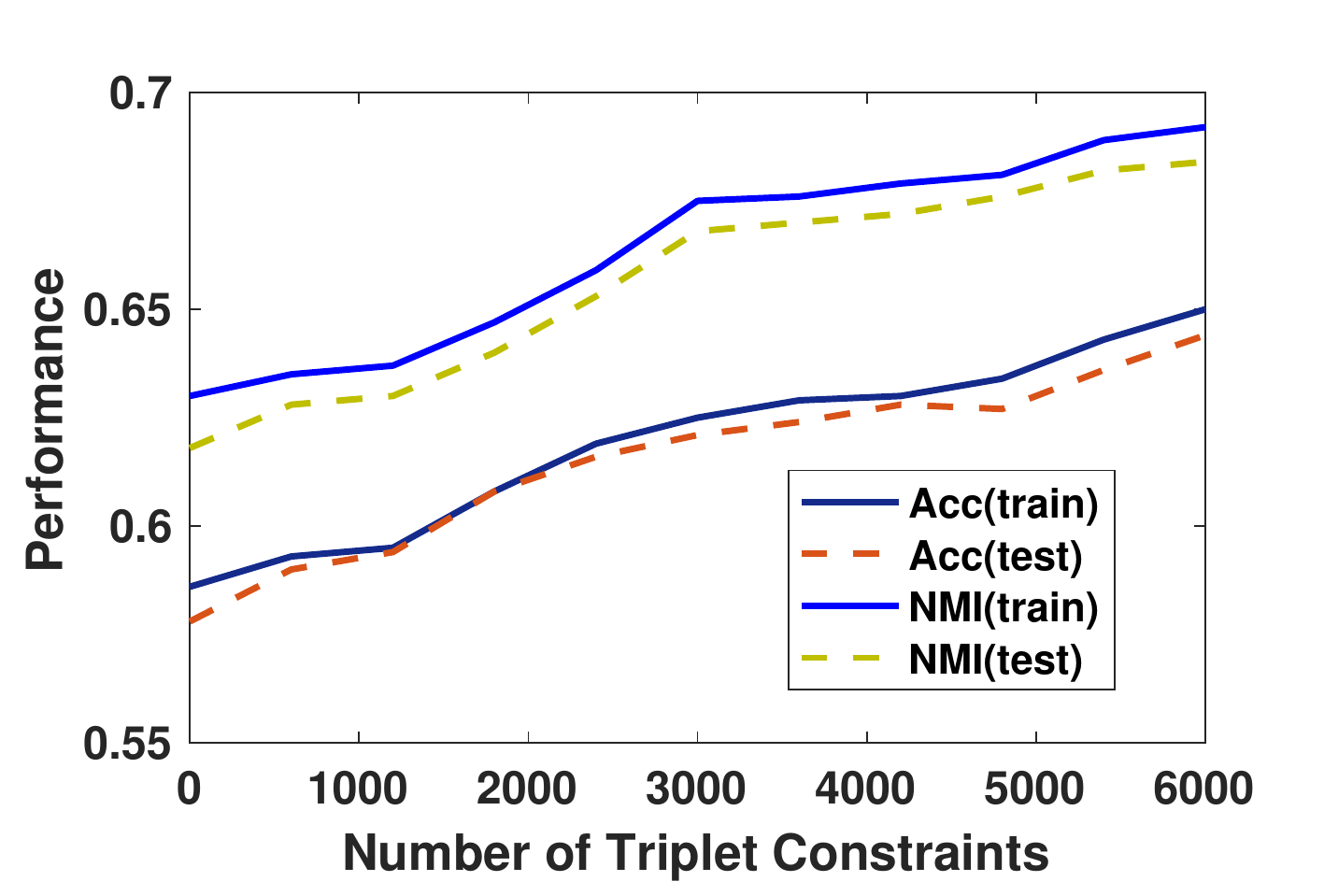}}
\caption{Evaluation of the effectiveness of triplet constraints in terms of Acc/NMI. }
\label{fig:triplet_performance}
\end{figure}

We experimented on MNIST and FASHION datasets. Figure \ref{fig:triplet_visual} visualizes example triplet constraints (based on embedding similarity), note the positive instances are closer to anchors than negative instances. In Figure \ref{fig:triplet_performance}, we show the clustering Acc/NMI improves consistently as the number of constraints increasing. Comparing with Figure \ref{fig:pairwise_performance}, we can find the pairwise constraints can bring slightly better improvements. That is because our triplet constraints are generated from a continuous domain and there is no exact together/apart information encoded in the constraints. Triplet constraints can be seen as a weaker but more general type of constraint.

\subsubsection{ Experiments on constraints generated from ontologies}
\label{exp:pairwise_triplet}
We experimented on the Fashion dataset. To show that pairwise constraints and triplet constraints generated from ontology can boost the performance with minimum supervision, we have randomly chosen $100$ training instances as a limited labeled set and generate full pairwise constraints based on these labeled instances. To generate the triplet constraints, we follow the procedure described in section \ref{sec:multiple_constraints}. Note the threshold for selecting positive pairs $\theta_{p}$ is set to be $0.5$ to ensure positive pairs are close and non-trivial (positive points are not all from the same classes), the threshold for negative pairs $\theta_{n}$ is set to be $0.3$ to be far away from anchors. We have generated $1000$ triplet constraints randomly from the same $100$ labeled training instances.

\begin{figure}[ht]
\centering
\subfigure[Fashion(init)]{\includegraphics[width=0.49\columnwidth]{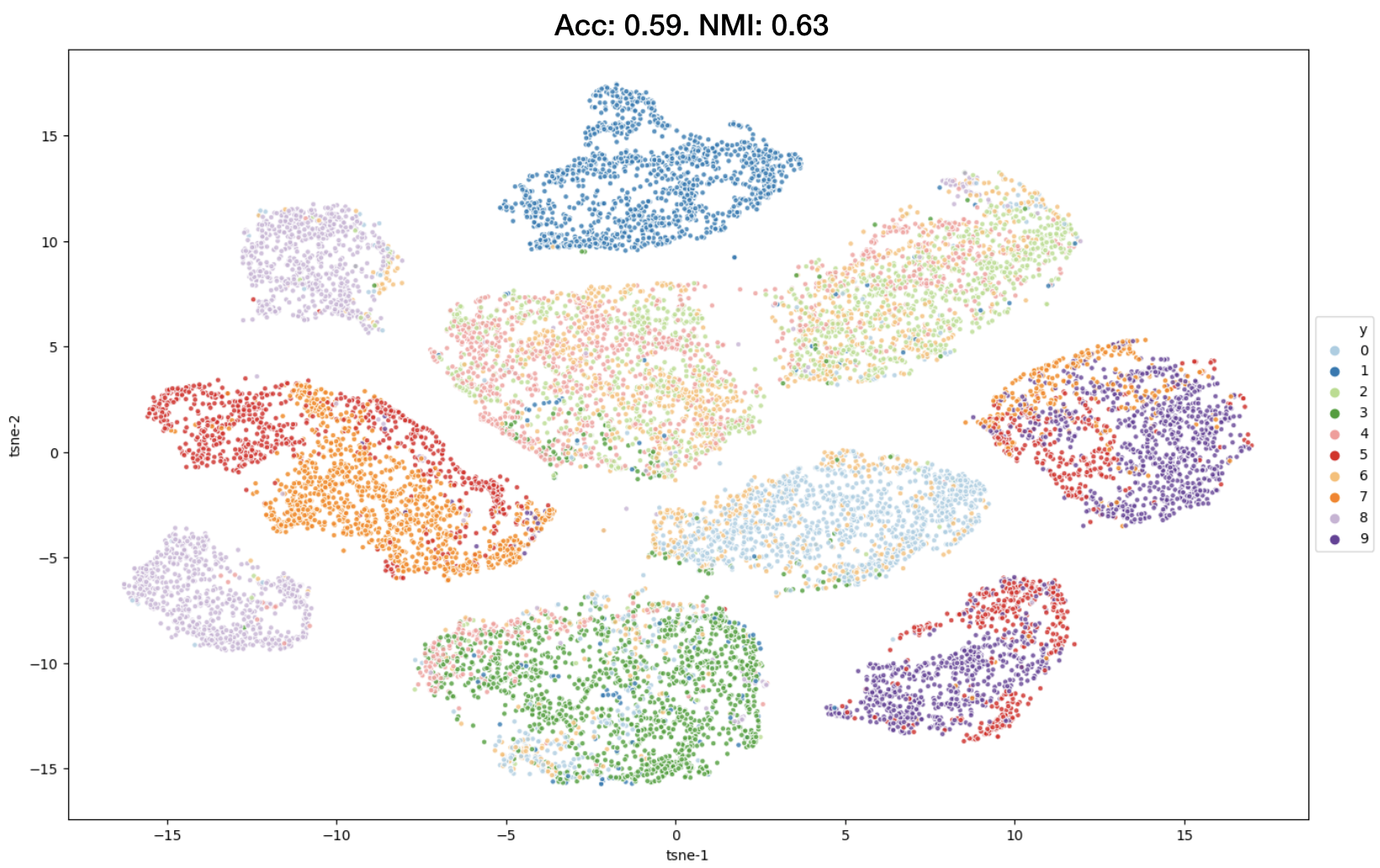}}
\hfill
\subfigure[Fashion(triplet)]{\includegraphics[width=0.49\columnwidth]{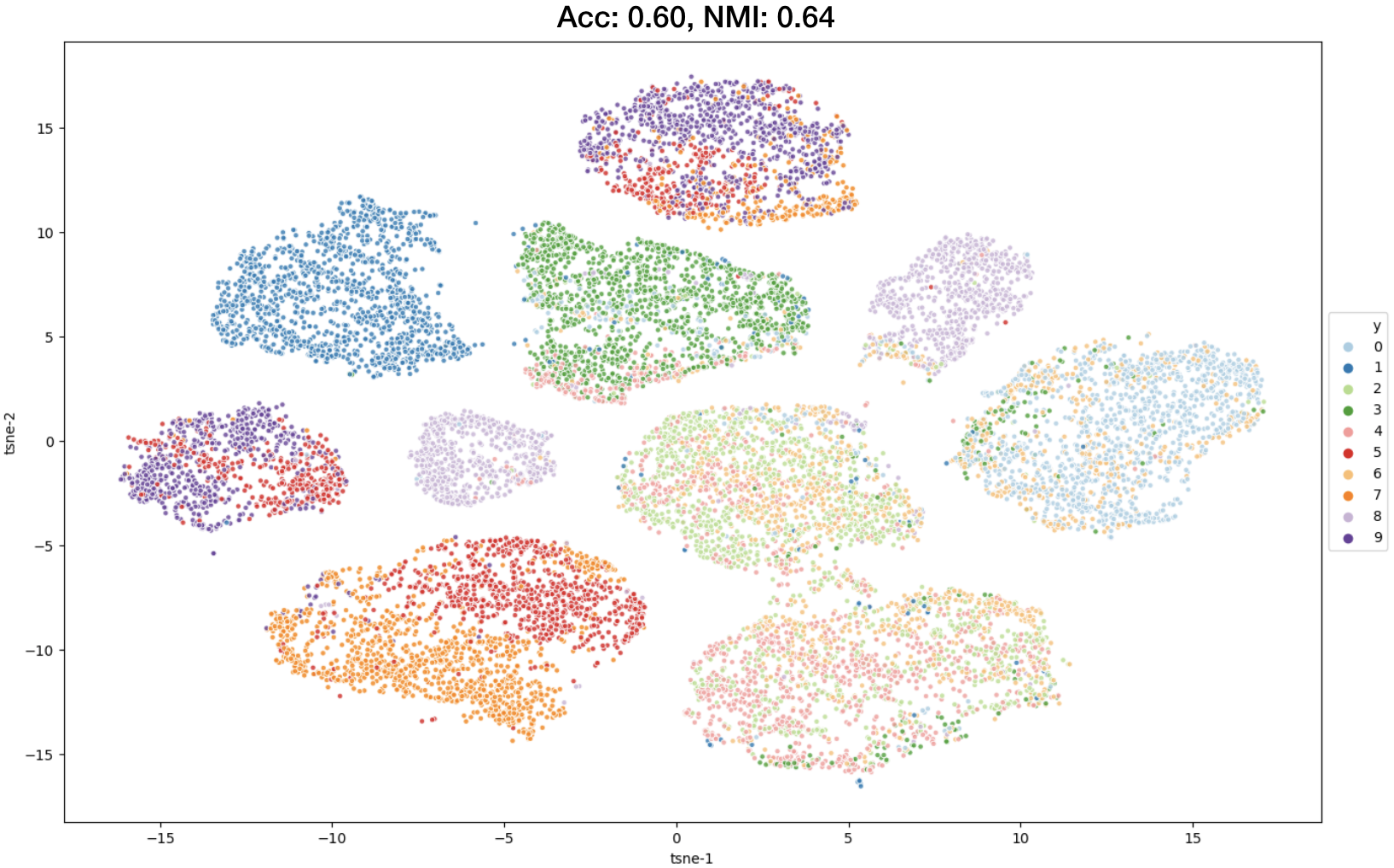}}
\subfigure[Fashion(pairwise)]{\includegraphics[width=0.49\columnwidth]{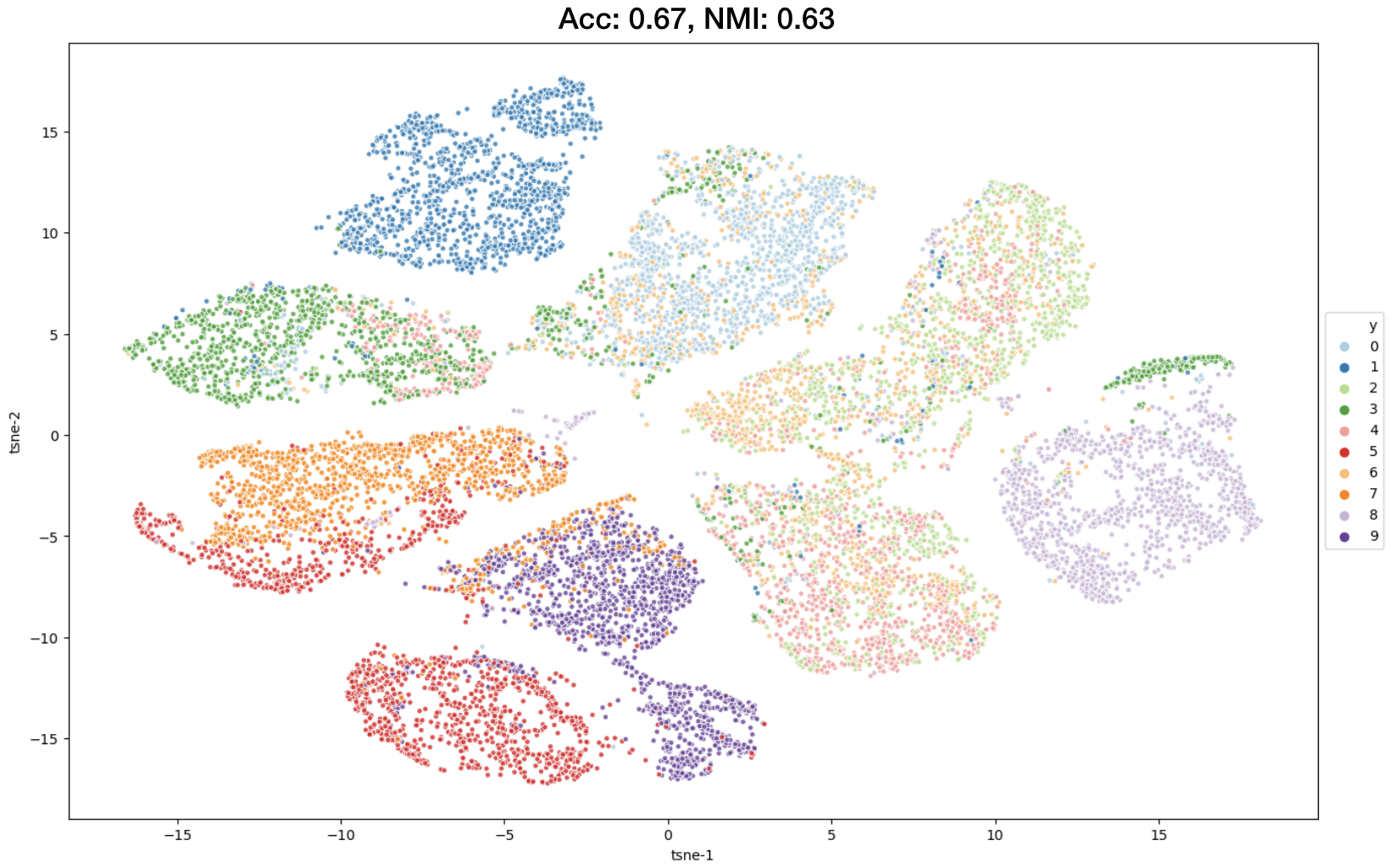}}
\hfill
\subfigure[Fashion(together)]{\includegraphics[width=0.486\columnwidth]{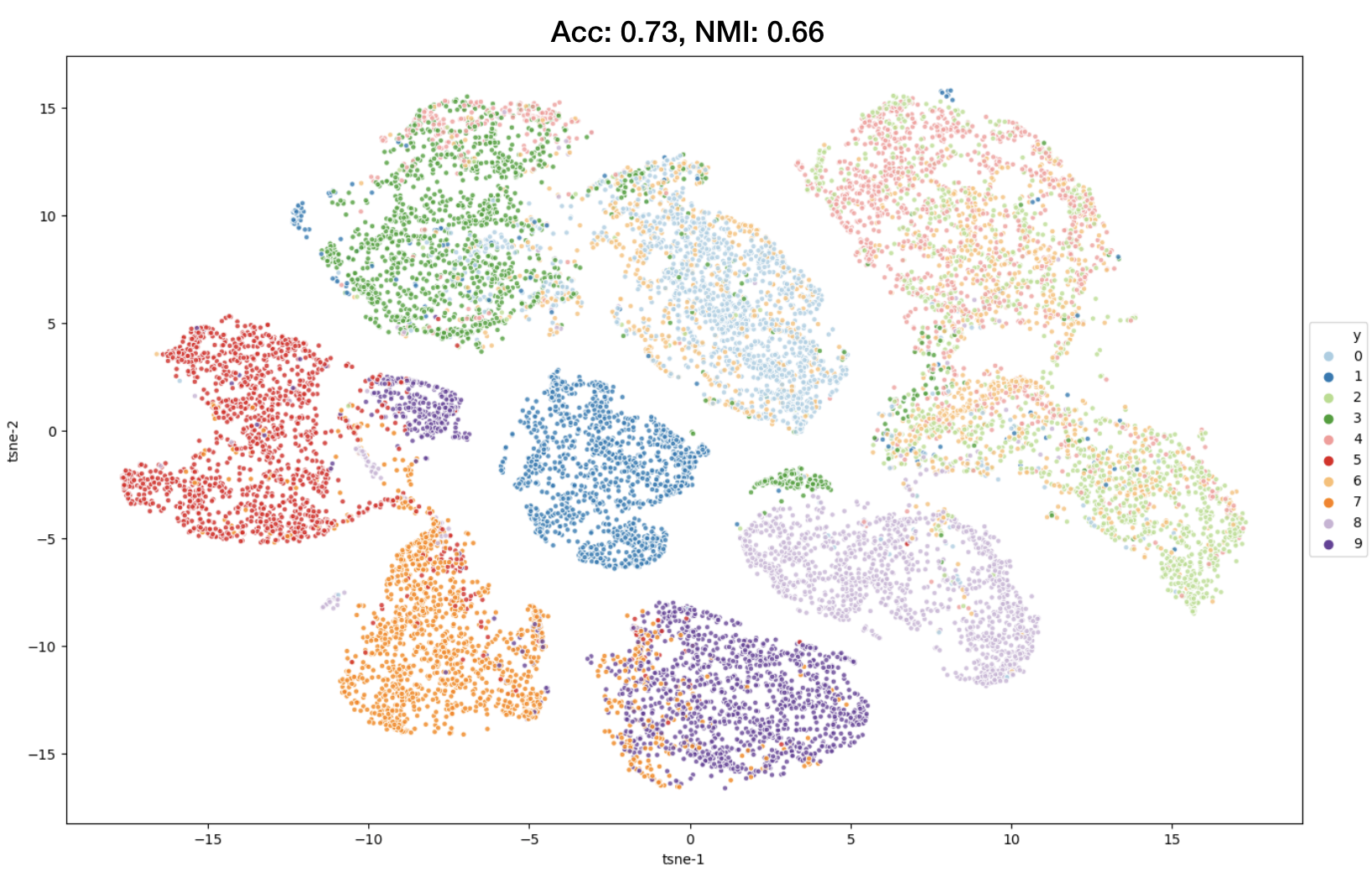}}

\caption{Experiments with Ontologies. Evaluation of the clustering performance of four settings. (a) No constraints, (b) Triplet constraints from labels, (c) Pairwise constraints from labels and (d) triplet constraints and pairwise constraints generated from WordNet ontology.}
\label{fig:pairwise_triplet}
\end{figure}

We empirically compare four different settings: i) clustering without any constraints, ii) clustering with just triplet constraints, iii) clustering with just pairwise constraints and iv) clustering with both pairwise and triplet constraints. Figure \ref{fig:pairwise_triplet} shows the embeddings we learned with four different settings with the corresponding clustering performance. We can see from the plots that both triplet constraints and pairwise constraints can improve the clustering performance when learned individually. Moreover, pairwise constraints can bring more significant improvement. The right bottom plot shows that learning with both these two types of constraints together can improve the clustering accuracy and clustering NMI in a large margin and achieve the highest performance. This shows that the triplet constraints generated from WordNet ontology can help regularize the latent space learned with pairwise constraints and yield a latent space which more similar to the ground truth.

\subsubsection{Experiments on global size constraints}
\label{exp:global}
\begin{figure}[ht]
\centering
\subfigure[MNIST]{\includegraphics[width=0.45\columnwidth]{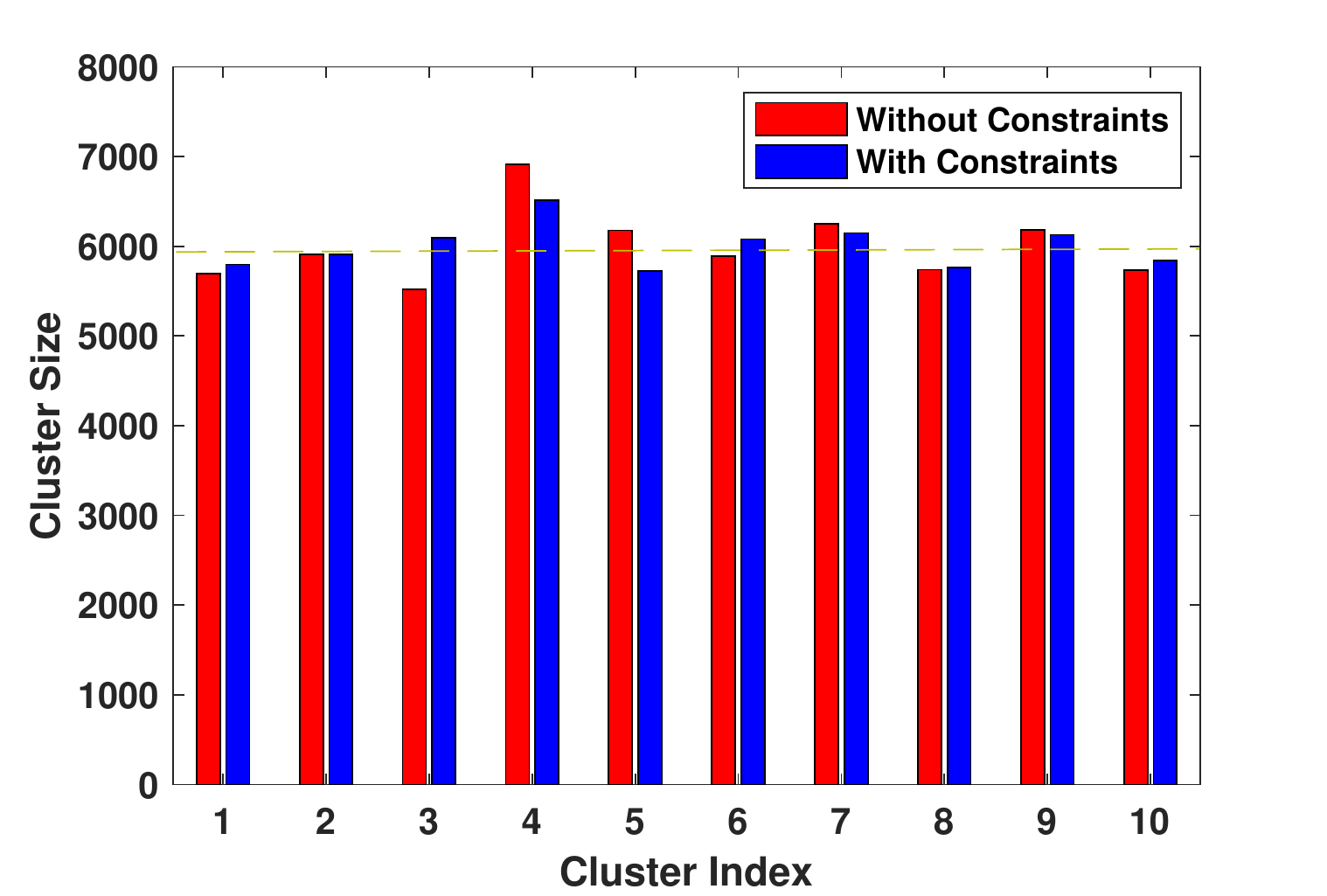}}
\hfill
\subfigure[Fashion]{\includegraphics[width=0.45\columnwidth]{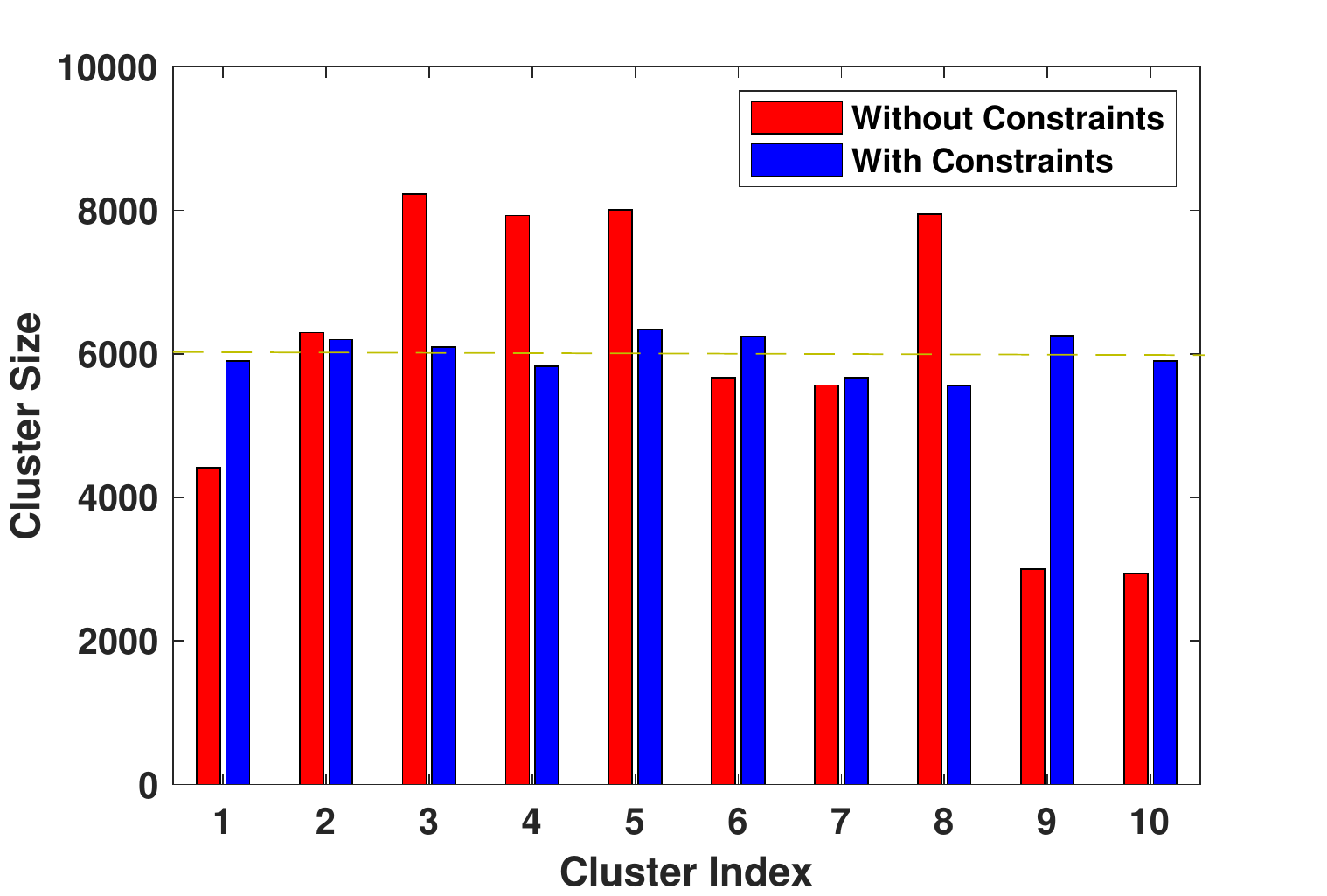}}
\caption{Evaluation of the global size constraints. This plot shows each cluster's size before/after adding global size constraints.}
\label{fig:global_constraints}
\end{figure}

To test the effectiveness of our proposed global size constraints, we have experimented on MNIST and Fashion training set since they both have balanced cluster sizes (see Figure \ref{fig:global_constraints}). Note that the ideal size for each cluster is $6000$ (each data set has $10$ classes); we can see that blue bars are more evenly distributed and closer to the ideal size.

We also evaluate the clustering performance with global constraints on MNIST (Acc:$0.91$, NMI:$0.86$) and Fashion (Acc:$0.57$, NMI:$0.59$). Comparing to the baselines in table \ref{tab:instance}, interestingly, we find the performance improved slightly on MNIST but dropped slightly on Fashion.

\subsubsection{ Experiments on noisy constraints}
\label{exp:robustness}
\begin{figure}[ht]
\centering
{\includegraphics[width=0.99\columnwidth]{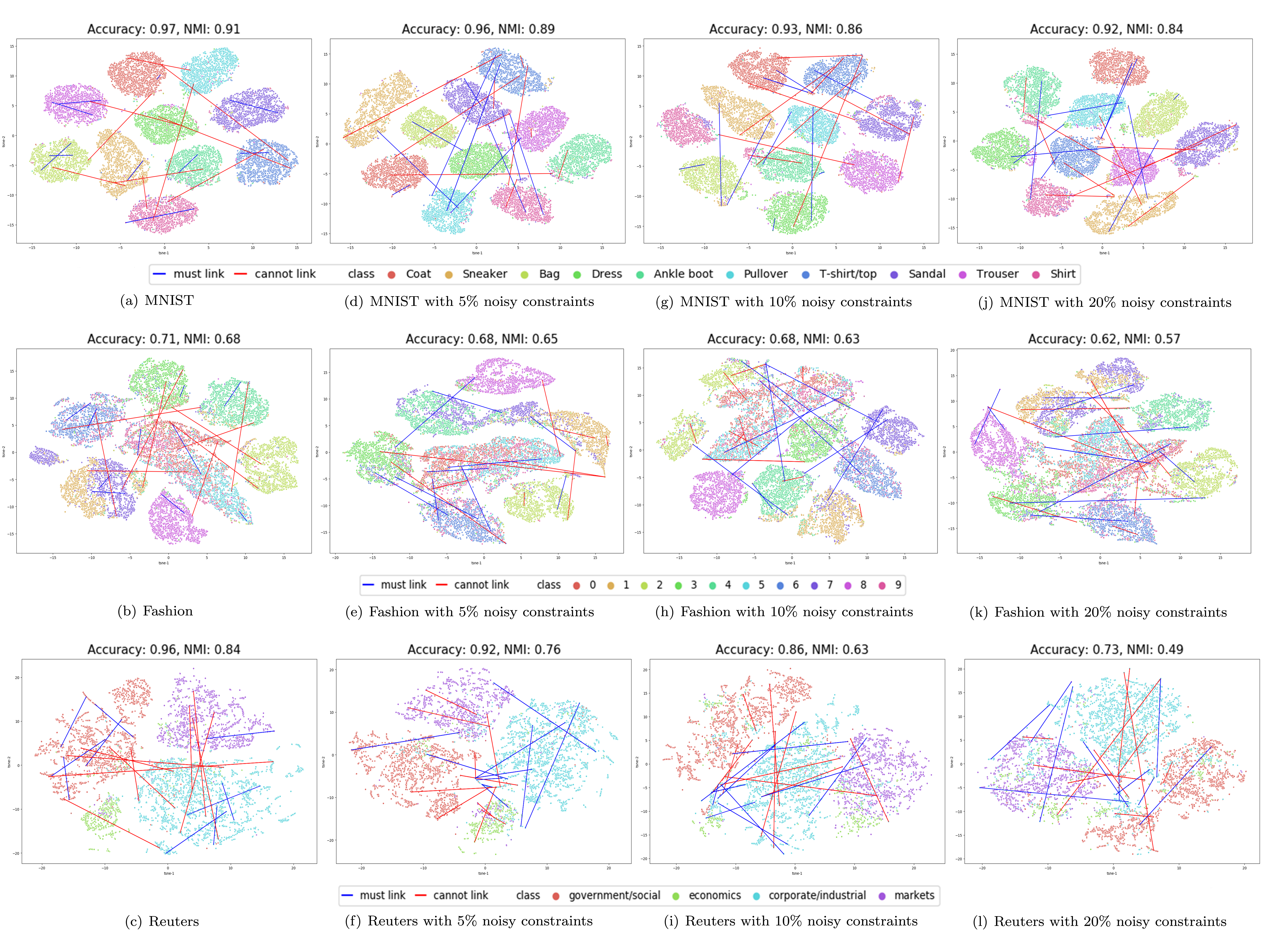}}
\caption{Effects of noisy constraints for MNIST, Fashion and Reuters Dataset.}
\label{fig:noisy_constraints}
\end{figure}

\textbf{Effect of Noisy Constraints.} To understand the effect of noisy constraints on our model, we randomly generate $6000$ pairs of constraints as described in Section \ref{exp:pairwise}. To generate noisy constraints, we first generate ground truth constraints and then flip the labels so that the true cannot-links become noisy must-links and the true must-links become noisy cannot-links. We define the degree of noisy constraints as the ratio of noisy constraints to ground truth constraints for each constraint type. For noisy degrees of 5\%, 10\%, 20\%, we randomly generated 300, 600, 1200 pairs of noisy constraints by flipping the labels of ground truth constraints. We visualized the embedded representation of a random subset of instances and its corresponding pairwise constraints using t-SNE and the learned embedding $z$. Figure \ref{fig:noisy_constraints} shows the cluster formation in training on MNIST, Fashion, Reuters dataset respectively.

We notice that the noisy constraint has negative effects on model performance. As the number of noisy constraints increases, the negative effect of noisy constraints on the model performance will also increase. For example, in Figure \ref{fig:noisy_constraints} the embedding without noisy constraints (plot (a)) has a better clustering result compared to the embedding with 20\% noisy constraints (plot (j)). Moreover, we notice that most of the noisy must-links are not satisfied and most of the noisy cannot-links are satisfied. To satisfy noisy cannot-links, the model will move instances from the correct cluster to another cluster that tends to have similar instances, which explains the negative effect noisy cannot-links have on the model performance. Figure \ref{fig:noisy_constraints}, plot (j) shows the model tries to satisfy some noisy cannot-links and forms a mixed cluster of instances ``3", ``5", and ``8". In the MNIST dataset, the instances with the label ``3", ``5", and label ``8" share some visual similarity. In plot (e, h, k), we observe a similar mixture of instances ``Sneaker," ``Sandal," ``Ankle boot." In the Fashion dataset, these classes all represent shoes and share some similarities.

\textbf{Robustness Against Noisy Constraints.} We define the noisy degree as the ratio of noisy constraints to ground truth constraints for one type of constraint. To test the model robustness against noisy constraints, we randomly generate $6000$ pairs constraints. For the noisy degree of 5\%, 10\%, 20\%, we randomly generate pairs of noisy constraints by flipping the labels of ground truth constraints and test the model performance. In each run, we fix the random seed and the initial centroids for k-means based methods. For each method, we compare its performance to the unconstrained version. In Table \ref{tab:noisy_constraintstab}, we show that on average, our model will start to perform worse than the unconstrained baseline model when the noisy degree in constraints reaches 20\%.
\begin{table}[h]
\tiny
\begin{center}
\caption{Pairwise constrained clustering performance (mean $\pm$ std) averaged over $50$ random noisy constraints sets. Baseline model is the model without using pairwise constraints.}
\begin{tabular}{cccccc}
\toprule[1.6pt]
Noise Degree& 0\% & 5\% & 10\% & 20\% & Baseline \\
\midrule
MNIST Acc & $0.962 \pm 0.01$& $0.953 \pm 0.01$ & $0.902 \pm 0.05$ &{$\textbf{0.883} \pm \textbf{0.05}$} & $0.883 \pm 0.01$ \\
MNIST NMI & $0.910 \pm 0.01$ & $0.894 \pm 0.02$ & $0.828 \pm 0.04$ &{$\textbf{0.809} \pm \textbf{0.04}$} & $0.861 \pm 0.01$ \\
\midrule
Fashion Acc & $0.737 \pm 0.04$ & $0.709 \pm 0.05$ & $0.695 \pm 0.04$ & $\textbf{0.681} \pm \textbf{0.05}$ & $0.587 \pm 0.01$ \\
Fashion NMI & $0.694 \pm 0.02$ & $0.666 \pm 0.03$ & $0.650 \pm 0.03$ & $\textbf{0.629} \pm \textbf{0.03}$ & $0.632 \pm 0.01$ \\
\midrule
Reuters Acc & $0.950 \pm 0.01$ & $0.856 \pm 0.20$ & $0.825 \pm 0.10$ & $\textbf{0.763} \pm \textbf{0.05}$ & $0.752 \pm 0.01$ \\
Reuters NMI & $0.818 \pm 0.01$ & $0.676 \pm 0.01$ & $0.578 \pm 0.01$ & $\textbf{0.503} \pm \textbf{0.04}$ & $0.542 \pm 0.02$ \\
\bottomrule[1.6pt]
\label{tab:noisy_constraintstab}
\end{tabular}
\end{center}
\end{table}

\subsubsection{Ablation Study}
\label{exp:ablation_study}
\textbf{Experiments on Initialization Approaches.}
To test the effect of different initialization approaches on our proposed deep clustering framework, we evaluate the model results for MNIST, Fashion, and Reuters dataset. Our model initializes both model weights and the cluster centers, so there are four initialization approaches. The ``Raw \& Rand" approach is to initialize both model weights and cluster centers randomly. ``Raw \& Kmeans" approach initializes cluster centers with KMeans and randomly initializes weights. The ``AE \& Rand" approach uses the pre-trained model to initialize weights and randomly initialize centroids. ``AE \& KMeans" uses Kmeans to initialize cluster centers and the pre-trained model to initialize model weights.
\begin{table}[ht]
\small
\begin{center}
\caption{Pairwise constrained clustering performance (mean $\pm$ std) averaged over $50$ random sets. Epoch 350*: model didn't converge after 350 epochs, where convergence is reached when the ratio of changed labels after an epoch $< 0.001$.}
\begin{tabular}{ccccc}
\toprule[1.6pt]
&Raw \& Rand & Raw \& KMeans & AE \& Rand & AE \& KMeans \\
\midrule
MNIST Acc & $0.880 \pm 0.07$& $0.915 \pm 0.06$ & $0.961 \pm 0.02$ &{$\textbf{0.962} \pm \textbf{0.01}$} \\
MNIST NMI & $0.830 \pm 0.06$& $0.859 \pm 0.05$ & $0.910 \pm 0.02$ &{$\textbf{0.910} \pm \textbf{0.01}$} \\
Epoch & $350$* & $350$* & $124.38 \pm 66.92$ &{$\textbf{107.60} \pm \textbf{35.62}$}\\
\midrule
Fashion Acc & $0.762 \pm 0.03$ & $0.757 \pm 0.03$ & $0.721 \pm 0.05$ &{$\textbf{0.737} \pm \textbf{0.04}$}\\
Fashion NMI & $0.697 \pm 0.01$ & $0.695 \pm 0.02$ & $0.680 \pm 0.03$ &{$\textbf{0.694} \pm \textbf{0.02}$}\\
Epoch & $350$* & $350$* & $350$* &{350*}\\
\midrule
Reuters Acc &$0.796 \pm 0.06$ & $0.797 \pm 0.06$ & $0.945 \pm 0.01$ &{$\textbf{0.950} \pm \textbf{0.01}$}\\
Reuters NMI &$0.585 \pm 0.08$ & $0.588 \pm 0.08$ & $0.809 \pm 0.02$ &{$\textbf{0.818} \pm \textbf{0.01}$}\\
Epoch & $47.73 \pm 16.37$ & $46.34 \pm 12.36$ & $9.33 \pm 4.34$ &{$\textbf{6.08} \pm \textbf{0.79}$}\\
\bottomrule[1.6pt]
\label{tab:initialization}
\end{tabular}
\end{center}
\end{table}

In Table \ref{tab:initialization}, we report the average performance with $6000$ randomly generated pairwise constraints. For MNIST and Reuters datasets, we compare the result for ``Raw \& Rand" with ``Raw \& Kmeans" and ``AE \& Rand" with ``AE \& Kmeans." We find that the cluster center initialization with Kmeans can increase training speed. We also observe that the consistent increase in model performance and training speed by comparing ``Raw \& Kmeans", ``Raw \& KMeans," ``AE \& Rand," and ``AE \& KMeans." This shows that better weight initialization can help the model learn information from pairwise constraints. However, for the Fashion dataset, the model performance becomes worse when using a pre-trained model to initialize weights. The result agrees with our findings in Section \ref{exp:pairwise}. This shows that the autoencoder's features are not always ideal for DEC's clustering objective. To address this issue, we can perform end-to-end deep constrained clustering from raw features.
\begin{table}[th]
\begin{center}
\caption{Ablation study to evaluate the contribution of clustering loss to pairwise constrained clustering. Note we report the mean clustering accuracy for each data set under two settings which $\ell_{C}$ means adding the clustering loss function.}
\begin{tabular}{cccccc}
\toprule[1.6pt]
& 600&1200 & 1800&2400&3000 \\
\midrule
MNIST Acc & $0.33$& $0.40$ & $0.45$ &$0.47$ & $0.49$ \\
MNIST Acc (with $\ell_{C}$) & $0.90$& $0.93$ & $0.95$ &$0.96$ & $0.97$ \\
\midrule
Fashion Acc & $0.52$& $0.56$ & $0.58$ &$0.60$ & $0.62$ \\
Fashion Acc (with $\ell_{C}$) & $0.59$& $0.61$ & $0.62$ &$0.63$ & $0.64$ \\
\midrule
Reuters Acc & $0.79$& $0.81$ & $0.83$ &$0.85$ & $0.86$ \\
Reuters Acc (with $\ell_{C}$) & $0.77$& $0.79$ & $0.81$ &$0.83$ & $0.85$ \\
\bottomrule[1.6pt]
\label{tab:clustering_loss}
\end{tabular}
\end{center}
\end{table}

\textbf{Evaluating the Contribution of Clustering Loss.}
To measure the contribution of clustering loss $\ell_{C}$ to our framework, we choose to study its influence on pairwise constrained clustering. We experiment on MNIST, Fashion, and Reuters data sets and report the average performance with a different number of randomly generated pairwise constraints. As shown in Table \ref{tab:clustering_loss}, the clustering loss is essential for image data sets, especially for the MNIST data set. The poor performance in MNIST has demonstrated the need to combine clustering loss with constraints learning. Otherwise, the network will overfit for a limited number of constraints.
Interestingly the results from the Reuters data set are opposite that adding clustering loss may harm the performance marginally. We hypothesis that the uni-model assumption, which encoded in the clustering loss function is preferred for image data rather than in text data. Another finding from the experimental results is that the performance gap between adding clustering loss or not is shrinking as the number of constraints increases. This is expected because as the number of constraints increasing the contribution of constraints loss is more and more critical.

\subsubsection{Experiments on Very Large Data Sets}
\label{exp:runtime_study}
Our previous experiments were on large data sets but under $100000$ instances, here we discuss these results and explore our method on a challenging real-world data set over $600000$ instances.
A key result of our results shown in Table \ref{tab:run_time} is that our method's increase in run-time over DEC is minimal for the three data sets previously studied. Interesting, our framework with instance-difficulty constraints is actually faster than the IDEC baseline, which speeds up the deep clustering procedure. We believe this is because this extra side information is compatible with the geometry of the data and hence increases converge to the minima. For the remaining three types of constraints, the running time is close to the IDEC's results.

\begin{table}[ht]
\begin{center}
\caption{Runtime analysis for our proposed approach with diffrent types of constraints. We use the same experimental setting for each type of constraints and average the running time (sec) over $10$ trails.}
\begin{tabular}{cccccc}
\toprule[1.6pt]
& IDEC & Pairwise & Instance & Global & Triplet \\
\midrule
MNIST& $178$& $197$ & $62$ &{$135$} & $230$ \\
\midrule
Fashion& $186$ & $246$ & $94$ & $217$ & $310$ \\
\midrule
Reuters & $5.15$ & $8.28$ & $3.82$ & $--$ & $--$ \\
\bottomrule[1.6pt]
\label{tab:run_time}
\end{tabular}
\end{center}
\end{table}

\begin{figure}[ht]
\centering
\subfigure[Clustering Performance on SVHN]{
\includegraphics[width=0.45\textwidth]{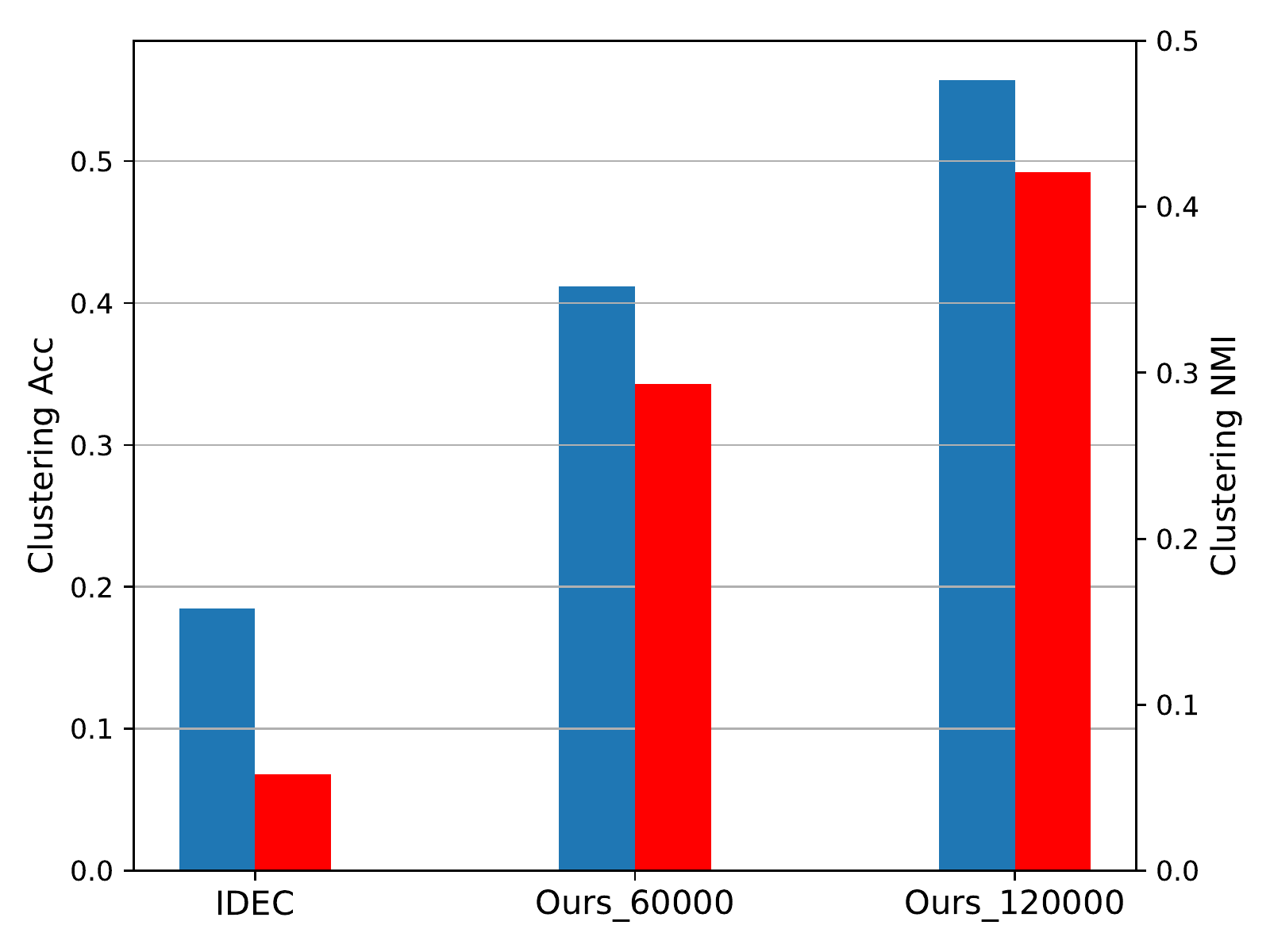}}
\hfill
\subfigure[Running Time on SVHN]{
\includegraphics[width=0.45\textwidth]{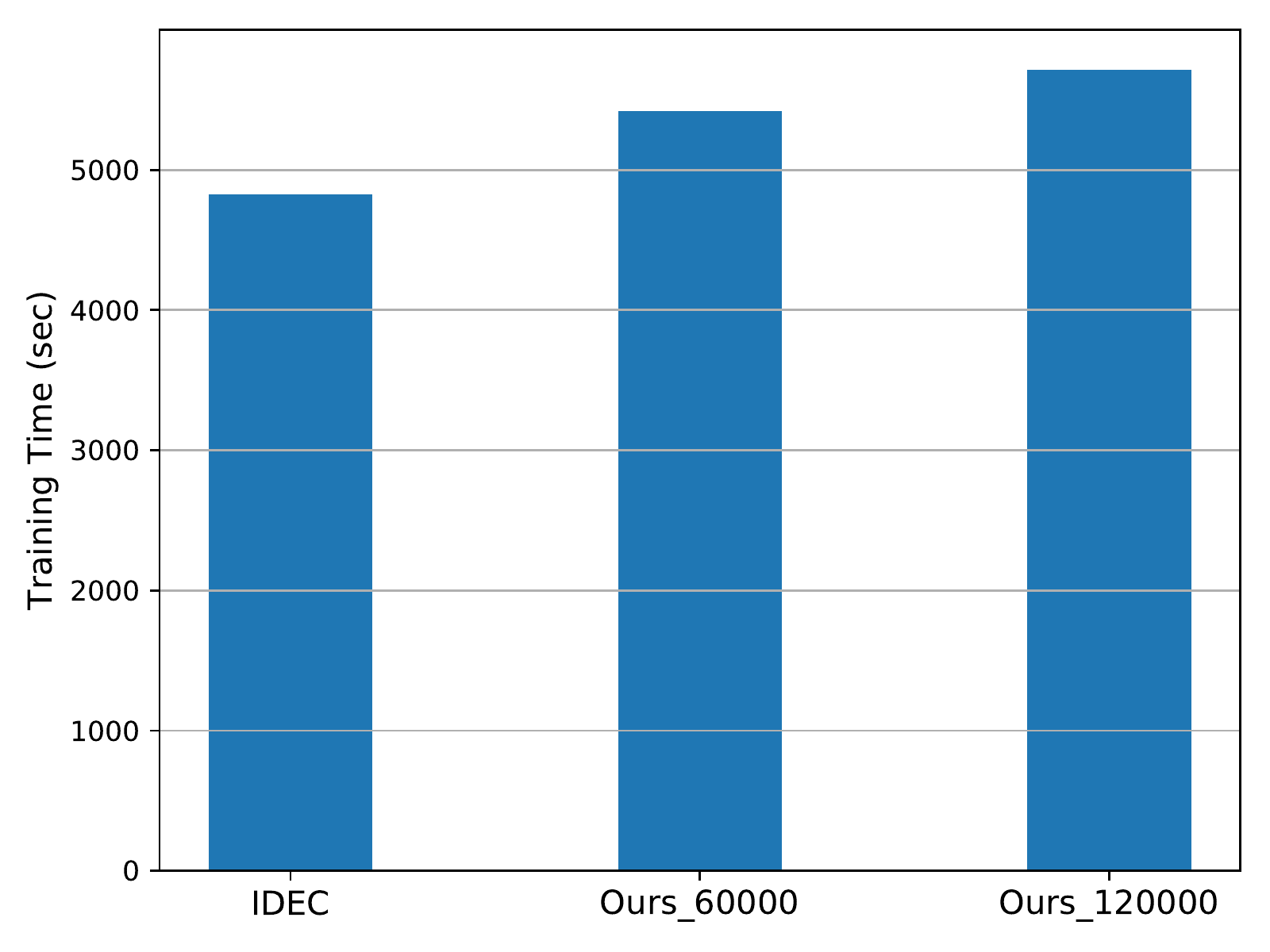}}
\caption{We report the out-of-sample prediction results of SVHN data in the left figure and the running time analysis in the right figure. Note we report the average clustering performance and running time (sec) over $10$ trails.}
\label{fig:svhn}
\end{figure}

We now study our method's run time on a very large data set, SVHN \citep{netzer2011reading}, which contains $604388$ training instances and $26032$ test instances. Compared to our previously used data (MNIST), this data set incorporates an order of magnitude more labeled data and comes from a significantly harder, unsolved, real-world problem (recognizing digits and numbers in natural scene images). 
We use the same experimental setting as our pairwise constrained clustering except that we generate more pairwise constraints ($60000$ and $120000$). We report the clustering performance as well as the time cost in Figure \ref{fig:svhn}. The clustering performance result is consistent with our earlier results and improves upon the accuracy of IDEC. Importantly, despite there being hundreds of thousands of constraints, the run time is only slightly more than the baseline IDEC algorithm. These running time results show that our approach is efficient and will not add too much overhead to deep clustering approaches.

\section{Conclusion, Limitations and Future Work}
\label{sec:limitations}
\label{sec:conclusion}

The area of constrained partitional clustering has a long history and is widely used. Constrained partitional clustering typically is mostly limited to simple pairwise together and apart constraints. In this paper, we show that deep clustering can be extended to a variety of fundamentally different constraint types, including instance-level (specifying hardness), cluster level (specifying cluster sizes), and triplet-level. We also show that our framework can not only handle standard constraints generated from labeled side information but new constraints generated from an ontology graph. Furthermore, we propose an efficient training paradigm that applies to multiple types of constraints simultaneously.

Our deep learning formulation was shown to advance the general field of constrained clustering in several ways. Firstly, it achieves better experimental performance than well-known k-means, mixture-model, and spectral constrained clustering in both an academic setting and a practical setting (see Table \ref{tab:pairwise_neg}).
Importantly, our approach does not suffer from the negative effects of constraints \citep{davidson2006measuring} as it learns a representation that simultaneously satisfies the constraints and finds a good clustering. This result is quite useful as a practitioner typically has just one constraint set, and our method is far more likely to perform better than using no constraints. Moreover, we have visualized our learning process to show how the learned latent representation overcomes inconsistencies and incoherence within the constraints.

Most constrained clustering approaches assume the oracle is perfect, and all the constraints are noise-free. Here we have also studied our model's robustness against noise in constraints, particularly the popular pairwise constraints. The experimental results demonstrate that our model is quite robust (see section \ref{exp:robustness}).
We were able to show that our method achieves all of the above but still retains the benefits of deep learning, such as scalability, out-of-sample predictions, and end-to-end learning. We found that even though standard non-deep learning methods were given the same representations (the auto-encoder embedding) of the data used to initialize our methods, the deep constrained clustering was able to adapt these representations even further.

Our current work limitations are two-folded: limitations inherent with the deep clustering backbone we have used (DEC/IDEC) and limitations with how we add constraints. In the first limitation, DEC or IDEC is doing k-means style clustering with limitations such as being partitional (i..e, no hierarchy and partial assignments). Moreover, the cluster number $k$ must be given apriori. As deep clustering evolves to more advanced styles of clustering, using the constraints we have explored in this paper seems reasonable. But the challenge of also having the deep learning solve for $k$ seems quite challenging given the need for a fixed architecture. As for the second limitation (how we add constraints), the main limitation is that we cannot solve for all constraint types at once without the need for multiple hyper-parameter tuning. This is part of a larger-scale problem in ML, how to tune hyperparameters (including $k$) efficiently.  We leave the current limitations as interesting future works.

\section*{Acknowledgments}
We acknowledge support for this work from a Google Gift entitled: ``Combining Symbolic Reasoning and Deep Learning".

\bibliographystyle{spbasic}
\bibliography{Manuscript}
\end{document}